\documentclass[12pt]{article}
\usepackage{amsmath}
\usepackage{amssymb}
\usepackage{graphicx}
\usepackage{float}
\usepackage{subfigure}
\usepackage{subfig}
\usepackage{makecell}
\usepackage{rotating}
\usepackage{adjustbox}
\usepackage{comment}

\newtheorem{rmk}{Remark}[section]

\usepackage{natbib}
\usepackage{url} 
\newcommand{\indep}{\perp \!\!\! \perp}
\newcommand\numberthis{\addtocounter{equation}{1}\tag{\theequation}}
\newcommand{\blind}{0}

\usepackage[dvipsnames, table]{xcolor}
\xdefinecolor{abricot}{named}{Apricot}
\xdefinecolor{limegreen}{named}{LimeGreen}
\xdefinecolor{Alizarin}{rgb}{0.82, 0.1, 0.26}

\begin{document}

\def\spacingset#1{\renewcommand{\baselinestretch}%
{#1}\small\normalsize} \spacingset{1}


\if0\blind
{
  \title{\bf Mixed Deep Gaussian Mixture Model: A clustering model for mixed datasets}
  \author{Robin Fuchs\footnote{robin.fuchs@univ-amu.fr}\\%
    CNRS, Centrale Marseille, I2M, MIO, Aix-Marseille Univ.\\
and \\
    Denys Pommeret
    \\
    Univ Lyon, UCBL, ISFA LSAF EA2429 \\
     and \\
    Cinzia Viroli \\
    Department of Statistical Sciences, Univ. of Bologna.
}
  \maketitle
} \fi

\if1\blind
{
  \bigskip
  \bigskip
  \bigskip
  \begin{center}
    {\LARGE\bf Title}
\end{center}
  \medskip
} \fi

\bigskip
\begin{abstract}
Clustering mixed data presents numerous challenges inherent to the very heterogeneous nature of the variables. A clustering algorithm should be able, despite of this heterogeneity, to extract discriminant pieces of information from the variables in order to design groups.
In this work we introduce a multilayer architecture model-based clustering method called Mixed Deep Gaussian Mixture Model (MDGMM) that can be viewed as an automatic way to merge the clustering performed separately on continuous and non-continuous data. This architecture is flexible and can be adapted to mixed as well as to continuous or non-continuous data. In this sense we generalize Generalized Linear Latent Variable Models and Deep Gaussian Mixture Models. We also design a new initialisation strategy and a data driven method that selects the best specification of the model and the optimal number of clusters for a given dataset ``on the fly". Besides, our model provides continuous low-dimensional representations of the data which can be a useful tool to visualize mixed datasets.
Finally, we validate the performance of our approach comparing its results with state-of-the-art mixed data clustering models over several commonly used datasets.
\end{abstract}

\noindent%
\vfill

\newpage
\spacingset{1.5} 
\section{Introduction}
Mixed data consist of variables of heterogeneous nature that can be divided into two categories: the continuous data generated by real-valued random variables, and the non-continuous data which are composed of  categorical and ordinal data (non-ordered or ordered data taking a finite number of modalities), binary data (that take either the value 1 or the value 0), and count data (taking values in $\mathbb{N}$).
By language abuse, these non-continuous variables will also be referred to as discrete variables in the following.

Due to their different natures, mixed variables do not share common scales and it is typically hard to measure the similarity between observations. 
There has been a significant and long interest in the statistical literature for mixed and continuous data clustering, which can be framed into four main categories, as described in  \cite{ahmad2019survey}: (i) partitional clustering minimizes the distance between observations and center groups by iterative optimization, as in K-modes or K-prototypes \citep{huang1997clustering,huang1998extensions}; (ii)
hierarchical algorithms perform nested clusterings and merge them to create the final clustering \citep{philip1983mixed,chiu2001robust}; (iii) model-based clustering \citep{mclachlan2000,fraley2002model,melnykov2010finite}, as their name suggests, rely on probability distributions; (iv) finally Neural Networks-based algorithms \citep{kohonen1990self} design the clusters thanks to connected neurons that learn complex patterns contained in the data. %

Within the framework of model-based clustering we propose a model for clustering mixed data, in which the different {non-continuous} variables are merged via a Generalized Linear Latent Variable Model (GLLVM) \citep{moustaki2003general,moustaki2000generalized}. GLLVMs assume that there exists a link function {between the non-continuous observable space (composed of non-continuous variables)} and a latent continuous data space, consisting of Gaussian latent variables. %
Recently, \cite{cagnone2014factor} have extended this approach by considering latent variables that are no more Gaussian but follow some mixtures of Gaussians \citep{fraley2002model} so as the observations are naturally clustered into groups. Since the latent dimension is chosen to be strictly lower than the original dimension, the model also performs dimension reduction. By abuse of language, we will refer to this extended version when mentioning GLLVMs in the sequel.

Our work generalizes this idea by considering a Deep Gaussian Mixture Model (DGMM) in the latent space \citep[see][]{viroli2019deep}. DGMMs can be seen as a series of nested
Mixture of Factor Analyzers (MFA) \citep{ghahramani1996algorithm,McLachlan2003}. As such, this approach performs clustering via subsequent dimensionally reduced latent spaces in a very flexible way.

To adapt the GLLVM to mixed data we propose a multilayer architecture inspired by the idea that composing simple functions enables to capture complex patterns, as in supervised neural networks. {We design two versions of our model. In the first one, denoted by M1DGMM, continuous and non-continuous data goes through the GLLVM model which acts as an embedding layer. The signal is then propagated to the following layers. In the second version, called M2DGMM, discrete data are still handled by the GLLVM model but continuous data are embedded separately by a DGMM head. The two signals are then merged by a ``common tail". This second architecture is analogous to multi-inputs Supervised Deep Learning architectures used for instance when data are composed of both images and text.}

Our model implementation relies on automatic differentiation \citep{baydin2017automatic} that helps keeping an acceptable running time even when the number of layers increases.
Indeed, using auto-differentiation methods provided for instance by the autograd package \citep{maclaurin2015autograd} cuts the computational running time. For instance, for the special case of GLLVM models, \cite{niku2019efficient} reported significant computational gains from using auto-differentiation methods.

To summarize, our work has three main aims:
it first extends the GLLVM and DGMM frameworks to deal with mixed data. Secondly, a new initialisation method is proposed to provide a suitable starting point for the MDGMM and more generally for GLLVM-based models. This initialization step combines Multiple Correspondence Analysis (MCA) or Factor analysis of mixed data (FAMD) which generalizes it, GMM,  MFA and the Partial Least Squares (PLS) algorithm. As mixed data are plunged into a multilayer continuous space we call this new initialisation Nested Spaces Embedding Procedure (NSEP). Thirdly, a model selection procedure is designed to identify the architecture of the model that best fits a given dataset.

Since the models are quite complex we propose to develop the method within the article and to reduce some mathematical developments by reporting them in a Supplementary Materials.

The paper is organized as follows:
Section \ref{model_prez} provides a detailed description of the proposed model. In Section \ref{model_training} the EM algorithms used for estimation are developed.
Section \ref{identif} deals with the identifiability constraints of the model. Section \ref{practical_consid} presents the initialization procedure NSEP and some practical considerations are given that can serve as a user manual. The performance of the model is compared to other competitor models in Section \ref{num_res}. In conclusion, Section \ref{conclusion} analyses the contributions of this work and highlights directions for future research.

\section{Model presentation}\label{model_prez}
\subsection{The MDGMM as a generalization of existing models}

{In the sequel we assume that we observe  $n$ random variables $y_1,\cdots, y_n$, such that  $\forall i =1,\cdots, n$, $y_i=(y^C_i, y^D_i)$, where $y_i^C$ is  a $p_C$-dimensional  vector of continuous random variables and $y_i^D$ is a $p_D$-dimensional vector of non-continuous random variables.  From what precedes, each $y_i$ is hence a vector of mixed variables of dimension $p=p_C+p_D$.}

The architecture of the MDGMM is based on  two models. First, Mixtures of Factor Analyzers generalized by the Deep Gaussian Mixture Models {are} applied on continuous variables, and second, a Generalized Linear Latent Variable Model {coupled with a DGMM} is applied on non-continuous variables. Mixtures of Factor Analyzers represent the most elementary building block of our model and can be formulated as follows:
\begin{equation*}
     y_i^C = \eta_{k} + \Lambda_{k} z_i + u_{i k}, \text{  with probability } \pi_{k},
\end{equation*}
where $k \in [1,K]$ identifies the group, $\eta_{k}$ is a constant vector of dimension $p_C$, $z_i \sim \mathrm{N}(0, I_r)$, $u_{i k} \sim \mathrm{N}(0, \Psi_k)$ and $\Lambda_{k}$ is the factor loading matrix of dimension $p_C \times r$, $r$ being the dimension of the latent space.
The underlying idea is to find a latent representation of the data of lower dimension $r$, with $r < p_C$. For each group $k$, the loading matrix is then used to interpret the relationship existing between the data and their new representation.

The DGMM approach consists in extending the MFA model by assuming that $z_i$ is no more drawn from a multivariate Gaussian but is itself a MFA. By repeating $L$ times this hypothesis we obtain a $L$-layers DGMM defined by:
\begin{equation}\label{DGMM}
    \begin{cases}
    y_i^C = \eta_{k_1}^{(1)} + \Lambda_{k_1}^{(1)} z_i^{(1)} + u_{i k_1}^{(1)},  \text{ with probability } \pi_{i, k_1}^{(1)} \\

    z_i^{(1)} = \eta_{k_2}^{(2)} + \Lambda_{k_2}^{(2)} z_i^{(2)} + u_{i k_2}^{(2)},  \text{ with probability } \pi_{i, k_2}^{(2)} \\
    \text{...} \\
    z_i^{(L-1)} = \eta_{k_{L}}^{(L)} + \Lambda_{k_{L}}^{(L)} z_i^{(L)} + u_{i k_L}^{(L)},  \text{ with probability } \pi_{i,k_L}^{(L)}  \\
    z_i^{(L)} \sim \mathcal{N}(0,I_{r_L}),
    \end{cases}
\end{equation}
where, for $\ell=1,\cdots, L$,  $k_{\ell} \in [1,K_{\ell}]$, $u_{ik_{\ell}}^{(\ell)}\sim N(0,\Psi_{k_{\ell}}^{(\ell)})$, $z_i^{(L)} \sim \mathrm{N}(0, I_{r_{L}})$  and where the factor loading matrices $\Lambda^{(\ell)}_{k_{\ell}}$ have dimension  $r_{\ell-1}\times r_{\ell}$, with the constraint $p > r_1 > r_2 > ... > r_L$.
Identifiability constraints on the parameters
$\Lambda_{k_{\ell}}^{(\ell)}$ and  $\Psi_{k_{\ell}}^{(\ell)}$ will be discussed in Section \ref{identif}.

The DGMM described in (\ref{DGMM}) can only handle continuous data. In order to apply a DGMM to discrete data we propose to integrate a Generalized Linear Latent Variable Model (GLLVM) framework within (\ref{DGMM}). {This new integrated model will be called Discrete DGMM (DDGMM)}.

A GLLVM assumes that, $\forall{j} \in [1,p_D]$, the discrete random variables $y_j^D$ are associated to one (or more) continuous latent variable through an exponential family link (see the illustrations given in \cite{cagnone2014factor}), under the so-called \emph{conditional independence assumption}, according to which variables are mutually independent conditionally to the latent variables.

Hence, one can combine the previously introduced DGMM architecture and the GLLVM to deal with mixed data.
In order to do so, we propose two specifications of the MDGMM: a one head version (the M1DGMM) and a two heads version (the M2DGMM). In the M1DGMM, the continuous variables pass through the GLLVM layer by defining a link function between $y^C$ and $z^{(1)}$ and one assumes that the \textit{conditional independence assumption} evoked earlier holds. On the contrary, by specifying a second head to deal with the continuous data specifically, one can relax this assumption: the continuous variables are not assumed to be mutually independent with respect to the latent variables. Instead, each continuous variable is only conditionally independent from the discrete variables but not from the other continuous variables. The two-heads architecture is also more flexible than the one-head specification as its ``link function" between $y^C$ and $z^C$ is a mixture of mixture rather than a probability distribution belonging to an exponential family. This flexibility comes at the price of additional model complexity and computational costs which has to be evaluated in regard of the performances of each specification.

The intuition behind the M2DGMM is simple.
The two heads extract features from the data and pass them to the common tail. The tail reconciles both information sources, designs common features and performs the clustering. As such, any layer on the tail could in principle be used as clustering layer. As detailed in Section \ref{model_selection_prez}, one could even use several tail layers to perform several clustering procedures (with different latent dimensions or numbers of clusters) in the same model run. The same remarks applies for the hidden layers of the M1DGMM.

To summarize the different setups that can be handled by DGMM-based models:
 \begin{itemize}
 \item
Use the M1DGMM or the M2DGMM when data are mixed,
\item
Use the DDGMM when data are non-continuous,
\item
Use the DGMM when data are continuous.
\end{itemize}

\subsection{Formal definition}
\begin{figure*}[h]%
\centering
\subfigure[M1DGMM]{\includegraphics[width=1.0\textwidth]{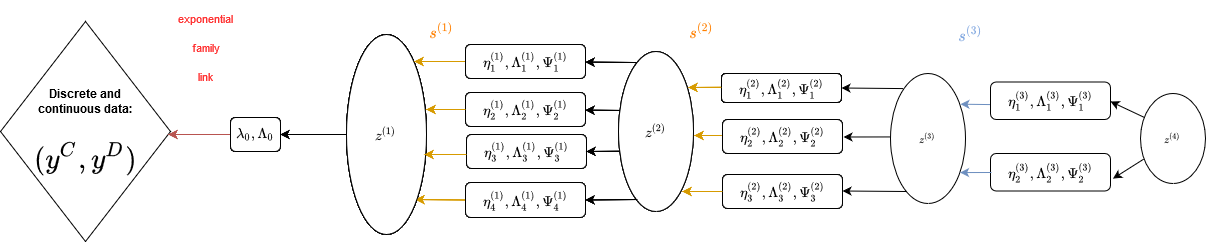}}%
\\
\subfigure[M2DGMM]{\includegraphics[width=1.0\textwidth]{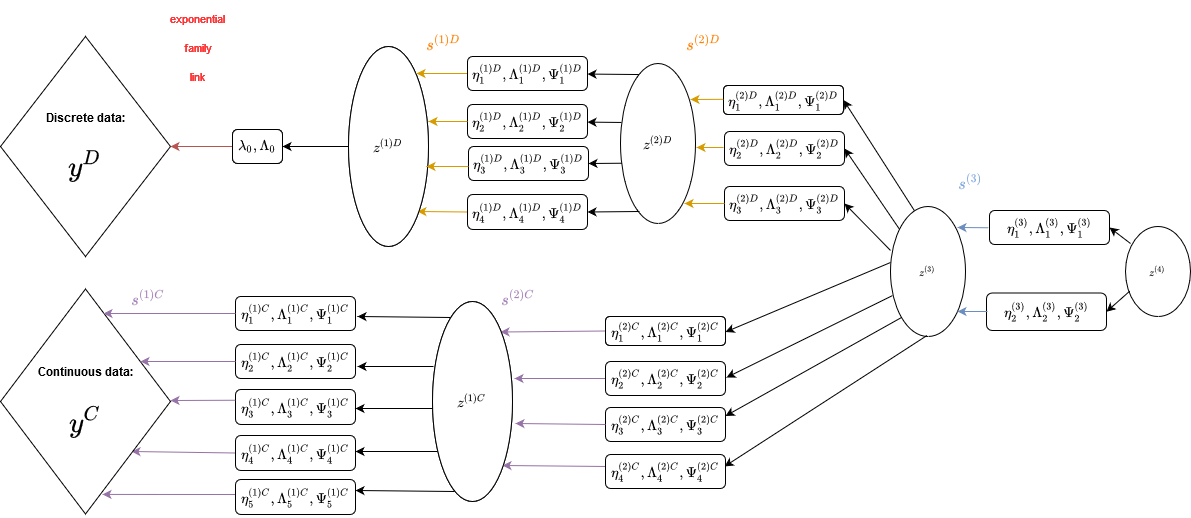}}%
\caption{Graphical model of: (a) M1DGMM, (b) M2DGMM\label{graph_model}}%
\end{figure*}

Let $y$ be the $n\times p$ matrix of the observed variables. We will denote by $i \in [1,n]$ the observation index and by $j \in [1,p]$ the variable index.
We can decompose the data as $y =( y^C , y^D)$ where $y^C$ is the  $n \times {p_C}$ matrix of continuous variables and  $y^D$ is the  $n \times {p_D}$ matrix of discrete variables. 
The global architecture of the M2DGMM is analogous to (\ref{DGMM}) with an additional GLLVM step for the discrete head as follows:
\begingroup
\allowdisplaybreaks
\begin{align}\label{MDGMM}
\text{Discrete head}:
    &\begin{cases}
    y_i^D \rightarrow z_i^{(1)D} \text{ through GLLVM link via }  (\lambda^{(0)}, \Lambda^{(0)}) \\
    z_i^{(1)D} = \eta_{k_1}^{(1)D} + \Lambda_{k_1}^{(1)D} z_i^{(2)D} + u_{i, k_1}^{(1)D}  \text{ with probability } \pi_{i,k_1}^{(1)D} \\
    \text{...} \\
    z_i^{({L_D})D} = \eta_{k_{{L_D}}}^{({L_D})D} + \Lambda_{k_{{L_D}}}^{({L_D})D} z_i^{({L_D} + 1)} + u_{k_{{L_D}}}^{({L_D})D},  \text{ with probability } \pi_{i,k_{{L_D}}}^{({L_D})D}
    \end{cases}\nonumber\\%
    \vspace{3mm}
\text{Continuous head}:
    &\begin{cases}
    y_i^C = \eta_{k_1}^{(1)C} + \Lambda_{k_1}^{(1)C} z_i^{(1)C} + u_{i, k_1}^{(1)C}  \text{ with probability } \pi_{i,k_1}^{(1)C} \\
    z_i^{(1)C} = \eta_{k_1}^{(1)C} + \Lambda_{k_1}^{(1)C} z_i^{(2)C} + u_{i, k_1}^{(1)C}  \text{ with probability } \pi_{i, k_2}^{(2)C} \\
    \text{...} \\
    z_i^{({L_C})C} = \eta_{k_{{L_C}}}^{({L_C})C} + \Lambda_{k_{{L_C}}}^{({L_C})C} z_i^{({L_C} + 1)} + u_{k_{{L_C}}}^{({L_C})},  \text{ with probability } \pi_{i,k_{{L_C} + 1}}^{({L_C} + 1)C}
    \end{cases}\\%
\text{Common tail}:
 &\begin{cases}
    z_i^{({L_0} + 1)} = \eta_{k_{{L_0}+1}}^{({L_0}+1)} + \Lambda_{k_{{L_0}+1}}^{({L_0}+1)} z_i^{({L_0}+2)} + u_{k_{{L_0}+1}}^{({L_0}+1)},  \text{ with probability } \pi_{i, k_{{L_0+2}}}^{({L_0}+1)} \\
    \text{...} \\
    z_i^{(L - 1)} = \eta_{k_{L - 1}}^{(L - 1)} + \Lambda_{k_{L-1}}^{(L-1)} z_i^{(L)} + u_{k_{L-1}}^{(L-1)}  \text{ with probability } \pi_{i,k_{L}}^{(L-1)} \\
    z_i^{(L)} \sim \mathcal{N}(0,I_{r_L}).\nonumber
\end{cases}
\end{align}
\endgroup
{The architecture of the M1DGMM is the same except that there is no ``continuous head" and that the $y_i^C$ goes through the GLLVM link. Figure \ref{graph_model} presents the graphical models associated with both specifications. In the M2DGMM case} one can observe
that $L_0= \max(L_C,L_D)$, that is, the first layer of the common tail is the $L_0+1$-th layers of the model.
For simplicity  of notation, we assume in the sequel that
$$L_C=L_D=L_0,$$
but all the results are easily obtained in the general case.
It is assumed that the random variables $(u_{k_{\ell}}^{(\ell)})_{k_{\ell}, \ell}$ are all independent.
The two heads only differ from each other by the fact that for the discrete head, a continuous representation of the data has first to be determined before information is fed through the layers. The GLLVM layer is parametrized by $(\lambda_{0}, \Lambda_0)$.
$\lambda_{0} = (\lambda_{0 bin}, \lambda_{0 count}, \lambda_{0 ord}, \lambda_{0 categ})$ contains the intercept coefficients for each discrete data sub-type. $\Lambda_{0}$ is a matrix of size $p_{D} \times r_1$, with $r_1$ the dimension of the first Discrete DGMM layer. 

The notation remains the same as in the previous subsection and only a superscript is added to specify for each variable the head or tail to which it belongs. For instance $z^C = (z^{(1)C}, ..., (z^{({L_C})C})$ is the set of latent variables of the continuous head. This subscript is omitted for the common head. The $\ell$-th layer of the head $h$  contains $K_{\ell}^h$ components which is the number of components of the associated  mixture. $L_D$ and  $L_C$ are the number of layers of the discrete and continuous head, respectively.

Each path from one layer to the next is the realization of a mixture.
In this sense we introduce, $s^{(\ell)h} \in [1,K_{\ell}^h]$ the latent variable associated with the index of the component $k_{\ell}^h$ of the layer ${\ell}$ of the head $h$. More generally, the latent variable associated with a path going from the first layer  to the last layer of one head $h$ is denoted by  $s^h = (s^{(1)h}, ..., s^{({L_0})h})$. Similarly, the random variable associated to a path going through all the common tail layers is denoted by $s^{({L_0}+1:)} = (s^{({L_0}+1)}, ..., s^{(L)})$. 
By extension, the variable associated with a full path going from the beginning of head $h$ to the end of the common tail is $s^{(1h: L)} = (s^h, s^{{L_0}+1:})$.  $s^{(1h: L)}$ belongs to $\Omega^h$ the set of all possible paths starting from one head of cardinal $S^h = \prod_{\ell=1}^L K_{\ell}^h$. The variable associated with a path going from layer $\ell$ of head $h$ to layer $L$ will be denoted $s^{(\ell h:L)}$. Finally, by a slight abuse of notation a full path going through the component $k_{\ell}^h$ of the $\ell$-th layer of head $h$ will be denoted: $s^{(1:k_{\ell}^h:L)}$ or more simply $s^{(:k_{\ell}^h:)}$. 

In order to be as concise as possible, we group the parameters of the model by defining:
\begin{align*}
    \Theta_D &= (\Theta_{emb}, \Theta_{DGMM}) = \left((\lambda_{0}, \Lambda_{0}),  (\eta_{k_{\ell}}^{(\ell)D}, \Lambda_{k_{\ell}}^{(\ell)D}, \Psi_{k_{\ell}}^{(\ell)D})_{k_{\ell} \in [1,K_{\ell}^D], \ell \in [1,{L_0}]}\right),\\
    \Theta_C &= (\eta_{k_{\ell}}^{(\ell)C}, \Lambda_{k_{\ell}}^{(\ell)C}, \Psi_{k_{\ell}}^{(\ell)C})_{k_{\ell} \in [1,K_{\ell}^C], \ell \in [1,{L_0}],} \ \ \ \
    \Theta_{{L_0}+1:} = (\eta_{k_{\ell}}^{(\ell)}, \Lambda_{k_{\ell}}^{(\ell)}, \Psi_{k_{\ell}}^{(\ell)})_{k_{\ell} \in [1,K_{\ell}], \ell \in [{L_0} + 1,L],}
\end{align*}
with $emb$ standing for embedding.

As an illustration, Figure \ref{graph_model} gives an example   graphical {models  for  M1DGMM and M2DGMM.}

{In Figure \ref{graph_model}, for the M2DGMM  case we have $K_C = (5,4)$, $K_D=(4, 3)$, $K = (2,1)$, $L_C=L_D={L_0} = 2$, $S^C = 40$ and $S^D = 24$. The decreasing size of the $(z^{(\ell)})_{\ell}$ illustrates the decreasing dimensions of the latent variables.}


\section{Model estimation}\label{model_training}
We deal only with the M2DGMM, the M1DGMM may be handled in much the same way.
The complete density of the M2DGMM is given by:
\begin{align*}
    L(y^C, y^D, & z^{C}, z^{D}, z^{({L_0}+1:)}, s^C, s^D,  s^{({L_0}+1:)} | \Theta_C, \Theta_D, \Theta_{{L_0}+1:})\\
    &= L(y^C | z^{(1)C}, s^C, s^{({L_0}+1:)}, \Theta_C, \Theta_{{L_0}+1:}) L(z^{C} | z^{({L_0}+1:)}, s^C, s^{({L_0}+1:)}, \Theta_C, \Theta_{{L_0}+1:})\\
    &\times L(y^D | z^{(1)D}, s^D, s^{({L_0}+1:)}, \Theta_D, \Theta_{{L_0}+1:})  L(z^{D} | z^{({L_0}+1:)}, s^D,  s^{({L_0}+1:)}, \Theta_D, \Theta_{{L_0}+1:}) \\
    &\times L(z^{({L_0}+1:)} | s^C, s^D,  s^{({L_0}+1:)}, \Theta_C, \Theta_D, \Theta_{{L_0}+1:})L(s^C, s^D,  s^{({L_0}+1:)} | \Theta_C, \Theta_D, \Theta_{{L_0}+1:}),
\end{align*}
which comes from the fact that we assume the two heads of the model to be conditionally independent with respect to the tail layers. Moreover, the layers of both heads and tail share the Markov property derived from the graphical model: $(z^{(\ell)h} \indep z^{(\ell+2)h}, ..., z^{(L)h}) \Big| z^{(l+1)h}$, $\forall{h} \in \{C, D, ({L_0} + 1:)\}$. 

The aim of the training is to maximize the expected log-likelihood, i.e. to maximize:
\begin{align*}
&\mathbb{E}_{z^{C}, z^{D}, z^{({L_0} + 1:)},  s^C, s^D, s^{({L_0}+1:)} | y^C, y^D,  \hat{\Theta}_C, \hat{\Theta}_D, \hat{\Theta}_{{L_0}+1:}}[\log L(y^C, y^D, z^{C}, z^{D}, z^{({L_0} + 1:)},  s^C, s^D, s^{({L_0}+1:)} | \Theta_C, \Theta_D, \Theta_{{L_0} + 1:})]
\end{align*}
that we derive in the Supplementary Materials.\\%
The model is fitted using a Monte Carlo version of the EM algorithm (MCEM) introduced by \cite{wei1990monte}. Three types of layers have here to be trained: the GLLVM layer, the regular DGMM layers and the common tail layers that join the two heads.

\subsection{Generalized Linear Latent Variable Model Embedding Layer}
In this section we present the canonical framework of GLLVMs for discrete data based on  \citet{moustaki2003general} and \citet{moustaki2000generalized}.\\
By the conditional independence assumption between discrete variables, the likelihood  can be written as:
\begin{align}
f(y^D | \Theta_D, \Theta_{{L_0} + 1:}) &= \int_{z^{(1)D}} \prod_{j = 1}^{p_D} f(y_j^{D} | z^{(1)D}, \Theta_D, \Theta_{{L_0} + 1:})f(z^{(1)D} | \Theta_D, \Theta_{{L_0} + 1:}) dz^{(1)D}, \label{gllvm_distrib}
\end{align}
where $y_j^D$ is the $j$th component of $y^D$. 
The density $f(y_j^{D} | z^{(1)D}, \Theta_D, \Theta_{{L_0} + 1:})$ belongs to an exponential family and in our empirical study we chose a Bernoulli distribution for binary variables, a binomial distribution for count variables and an ordered multinomial distribution for ordinal data. The whole expressions of the densities %
can be found in \cite{cagnone2014factor}. 
In order to train the GLLVM layer, we  maximize
\begin{align*}
& \mathbb{E}_{z^{(1)D}, s^D, s^{({L_0} + 1:)} | y^D, \hat{\Theta}_D, \hat{\Theta}_{{L_0} + 1:}}[\log L(y^D| z^{(1)D}, s^D, s^{{L_0} + 1:}, \Theta_D, \Theta_{{L_0} + 1:})] \\
&= \mathbb{E}_{z^{(1)D}| y^D, \hat{\Theta}_D, \hat{\Theta}_{{L_0} + 1:}}[\log L(y^D| z^{(1)D}, \Theta_D, \Theta_{{L_0} + 1:})]\\
&= \int f(z^{(1)D} | y^D, \hat{\Theta}_D, \hat{\Theta}_{{L_0} + 1:})\log L(y^D |z^{(1)D}, \Theta_D,  \Theta_{{L_0} + 1:})dz^{(1)D},
\end{align*}
the second equality being due to the fact that  $y^D$ is related to $(s^D, s^{({L_0}+1:)})$ only through $z^{(1)D}$.

\subsubsection{MC Step}
Draw $M^{(1)}$ observations from $f(z^{(1)D} | s^D, s^{({L_0}+1:)}, \hat{\Theta}_D, \hat{\Theta}_{{L_0} + 1:})$.

\subsubsection{E step}
Hence the E step consists in determining $f(z^{(1)D} | y^D, \hat{\Theta}_D, \hat{\Theta}_{{L_0} + 1:})$, which can be rewritten as:
\begin{align}
        f(z^{(1)D} | y^D, \hat{\Theta}_D, \hat{\Theta}_{{L_0} + 1:})  = \sum_{s'}f(z^{(1)D} | y^D, s', \hat{\Theta}_D, \hat{\Theta}_{{L_0} + 1:})f(s^{(1D:L)} = s'| y^D, \hat{\Theta}_D, \hat{\Theta}_{{L_0} + 1:}).
        \label{p(zs_y)}.
\end{align}
The detailed calculus is given in the Supplementary Materials.

\subsubsection{M step}
There are no closed-form solutions for the estimators of $(\lambda_0, \Lambda_0)$ that maximize
$$\mathbb{E}_{z^{(1)D}| y^D, \hat{\Theta}_D, \hat{\Theta}_{L_0 + 1:}}[\log L(y^D| z^{(1)D}, \Theta_D, \hat{\Theta}_{L_0 + 1:})].$$ We then  use optimisation methods (see Supplementary Materials).

\subsection{Determining the parameters of the DGMM layers}
In this section, we omit the subscript $h \in \{C, D\}$ on the $z^h$, $y^h$ and $s^h$ variables because the reasoning is the same for both cases.
For $\ell \in [1,{L_0}]$, we aim to maximize $$\mathbb{E}_{z^{(\ell)}, z^{(\ell+1)}, s | y,  \hat{\Theta}}[\log L(z^{(\ell)} | z^{(\ell+1)}, s, \Theta)].$$
Here the conditional distribution under which the expectation is taken depends on variables located in 3 different layers.

\subsubsection{MC Step}
At each layer $\ell$, $M^{(\ell)}$ pseudo-observations are drawn for each of the previously {obtained} $\prod_{j=1}^{\ell-1} M^{(j)}$ pseudo-observations.
Hence, in order to draw from $f(z^{(\ell)}, z^{(\ell+1)},  s | y,  \hat{\Theta})$ at layer $\ell$:
\begin{itemize}
    \item If $\ell=1$, reuse the $M^{(1)}$ pseudo-observations drawn from $f(z^{(1)} | s, \hat{\Theta})$,
    \item otherwise reuse the $ M^{(\ell-1)}$ pseudo-observations from $f(z^{(\ell-1)} | y, s, \hat{\Theta})$ and the $M^{(\ell)}$ pseudo-observations from $f(z^{(\ell)} | z^{(\ell-1)}, s, \hat{\Theta})$ computed for each pseudo-observation of the previous DGMM layer.
    \item Draw $M^{(\ell+1)}$ observations from $f(z^{({\ell}+1)} | z^{(\ell)}, s, \hat{\Theta})$ for each previously sampled $z^{(\ell)}$.
\end{itemize}

\subsubsection{E Step}
The conditional expectation distribution has the following decomposition:
\begin{align}
    f(z^{(\ell)}, z^{(\ell+1)}, s | y,  \hat{\Theta})
    &= f(z^{(\ell)}, s | y,  \hat{\Theta}) f(z^{(\ell+1)} | z^{(\ell)}, s, y, \hat{\Theta}) \nonumber \\
    &= f(z^{(\ell)} | y,  s, \hat{\Theta}) f(s | y, \hat{\Theta}) f(z^{(\ell+1)} | z^{(\ell)}, s, \hat{\Theta}),
    \label{E_distrib_l}
\end{align}
and we develop this term in the Supplementary Materials.

\subsubsection{M step}
The estimators of the DGMM layer parameters $\forall \ell \in [1, {L_0}]$ are given by:
$$
\begin{cases}
\displaystyle
\hat{\eta}_{k_{\ell}}^{(\ell)} = \frac{\sum_{i=1}^{n} \sum_{\tilde{s}_i^{(:k_{\ell}:)}} f(s_i^{(:k_{\ell}:)} = \tilde{s}_i^{(:k_{\ell}:)}|y, \hat{\Theta})\left[
E[z_i^{(\ell)} | s_i^{(:k_{\ell}:)} = \tilde{s}_i^{(:k_{\ell}:)}, y_i,  \hat{\Theta}] - \Lambda_{k_{\ell}}^{(\ell)}E[z_i^{(\ell+1)}| \tilde{s}_i^{(:k_{\ell}:)},  y_i,  \hat{\Theta}]\right]}{\sum_{i=1}^{n} \sum_{\tilde{s}_i^{(:k_{\ell}:)}} f(s_i^{(:k_{\ell}:)} = \tilde{s}_i^{(:k_{\ell}:)}|y_i, \hat{\Theta})} \\
\hat{\Lambda}_{k_{\ell}}^{(\ell)} =
\frac{\sum_{i=1}^n \sum_{\tilde{s}_i^{(:k_{\ell}:)}} f(s_i^{(:k_{\ell}:)} = \tilde{s}_i^{(:k_{\ell}:)}| y_i, \hat{\Theta}) \Big[E[(z_i^{({\ell})} - \hat{\eta}_{k_{\ell}}^{({\ell})})z_i^{({\ell}+1)T} | s_i^{(:k_{\ell}:)} = \tilde{s}_i^{(:k_{\ell}:)}, y_i,  \hat{\Theta}]\Big]}{\sum_{i=1}^n \sum_{\tilde{s}_i^{(:k_{\ell}:)}} f(s_i^{(:k_{\ell}:)} = \tilde{s}_i^{(:k_{\ell}:)}| y_i, \hat{\Theta})}E[z_i^{({\ell}+1)}z_i^{({\ell}+1)T} | \tilde{s}_i^{(:k_{\ell}:)}, y_i, \hat{\Theta}]^{-1}\\

\hat{\Psi}^{({\ell})}_{k_{{\ell}}} =
\frac{\sum_{i=1}^n \sum_{\tilde{s}_i} f(s_i^{(:k_{\ell}:)} = \tilde{s}_i^{(:k_{\ell}:)}| y_i, \hat{\Theta})E\Big[\Big{(}z_i^{({\ell})} - (\eta_{k_{l}}^{({\ell})} +  \Lambda_{k_{l}}^{({\ell})}z_i^{({\ell}+1)})\Big)\Big{(}z_i^{({\ell})} - (\eta_{k_{l}}^{({\ell})} +  \Lambda_{k_{l}}^{({\ell})}z_i^{({\ell}+1)})\Big)^T | \tilde{s}_i^{(:k_{\ell}:)}, y_i,  \hat{\Theta} \Big]}{\sum_{i=1}^n \sum_{\tilde{s}_i^{(:k_{\ell}:)}} f(s_i^{(:k_{\ell}:)} = \tilde{s}_i^{(:k_{\ell}:)}| y_i, \hat{\Theta})},   \\

\label{est}
\end{cases}
$$
with $\tilde{s}_i^{(:k_{\ell}:)} = (\tilde{k}_1, ..., \tilde{k}_{{\ell}-1}, k_{{\ell}}, \tilde{k}_{{\ell}+1} , ..., \tilde{k}_L)$, a path going through the network and reaching the component $k_{{\ell}}$.
The details of the computation are given in the Supplementary Materials. %

\subsection{Training of the common tail layers}
In this section we aim to maximise $\forall{{\ell}} \in [{L_0} + 1, L]$, the following expression:
$$\mathbb{E}_{z^{({\ell})}, z^{({\ell}+1)},  s^C, s^D, s^{({L_0}+1:)} | y^C, y^D,  \hat{\Theta}_C, \hat{\Theta}_D, \hat{\Theta}_{{L_0}+1:}}[\log L(z^{(\ell)} | z^{({\ell} + 1)}, s^C, s^D, s^{({L_0}+1:)}, \Theta_C, \Theta_D, \Theta_{{L_0}+1:})].$$

\subsubsection{MC Step}
The MC step remains the same as for regular DGMM layers except that the conditioning concerns both types of data ($y^C$ and $y^D$) and not only discrete or continuous data as in the heads layers.

\subsubsection{E Step}
The distribution of the conditional expectation is $f(z^{(\ell)}, z^{(\ell+1)}, s^C, s^D, s^{({L_0}+1:)} | y^C, y^D,  \hat{\Theta}_C, \hat{\Theta}_D, \hat{\Theta}_{{L_0}+1:})$ that we can express as previously.
We detail the calculus in the Supplementary Materials.

\subsubsection{M Step}
The estimators of the junction layers keep the same form as the regular DGMM layers except once again that the two types of data and paths exist in the conditional distribution of the expectation.

\subsection{Determining the path probabilities}\label{paths_proba_sec}
In this section, we determine the path probabilities by optimizing the parameters of the following expression derived from the expected log-likelihood: %
$$\mathbb{E}_{s^C, s^D, s^{({L_0}+1:)}|y^C, y^D, \hat{\Theta}_C, \hat{\Theta}_D, \hat{\Theta}_{{L_0}+1:}}[\log L(s^C, s^D, s^{({L_0} + 1:)} | \Theta_C, \Theta_D, \Theta_{{L_0} + 1:})],$$
with respect to $\pi_s^{h}$, $\forall{h \in \{C, D\}}$ and $\pi_s^{({L_0} + 1:)}$.

\subsubsection{E step}
By mutual independence of $s^C, s^D$ and $s^{{L_0} + 1:}$, estimating the distribution of the expectation boils down to computing three densities:  $f(s^{({\ell} )D} = k_{\ell}  | y^D, \hat{\Theta}_D, \hat{\Theta}_{{L_0}+1:})$, $f(s^{({\ell} )C} = k_{\ell}  | y^C, \hat{\Theta}_C, \hat{\Theta}_{{L_0}+1:})$, and $f(s^{({\ell} )} = k_{\ell}  | y^C, y^D, \hat{\Theta}_C, \hat{\Theta}_D, \hat{\Theta}_{{L_0}+1:})$ (details are given in the Supplementary Materials).

\subsubsection{M step}
Estimators for each head $h$ and for the common tail are given respectively by (see the Supplementary Materials):
\begin{align*}
\hat{\pi}_{k_{\ell} }^{({\ell} )h} = \frac{\sum_{i = 1}^n  f(s_i^{({\ell} )h} = k_{\ell}  | y_i^h, \hat{\Theta}_h, \hat{\Theta}_{{L_0}+1:})}{n}
& \quad {\rm and \ } \quad
 \hat{\pi}_{k_{\ell} }^{({\ell} )} = \frac{\sum_{i = 1}^n  f(s_i^{({\ell} )} = k_{\ell}  | y_i^C, y_i^D, \hat{\Theta}_C, \hat{\Theta}_D, \hat{\Theta}_{{L_0}+1:})}{n}.
\end{align*}

%

\section{Identifiability}\label{identif}
In this section, we combine both GLLVM and DGMM identifiability constraints proposed in \cite{cagnone2014factor} and \cite{viroli2019deep}, respectively, to make our model identifiable.

\subsection{GLLVM identifiability constraints}
{Both the GLLVM model and the Factor Analysis model assume that the latent variables are centered and of unit variance.
This can be obtained by rescaling iteratively all the latent layers parameters from the last common layer to the first head layers as follows: }

$$\begin{cases}
    \eta_{k_\ell}^{(\ell)h new} = (A^{(\ell)h})^{-1T}\left[\eta_{k_\ell}^{(\ell)h} - \sum_{k'_{\ell}} \pi_{k'_{\ell}}^{(\ell)h} \eta_{k'_{\ell}}^{(\ell)h} \right] \\
    \Lambda_{k_\ell}^{(\ell)h new} = (A^{(\ell)h})^{-1T}\Lambda_{k_\ell}^{(\ell)h} \\
    \Psi_{k_\ell}^{(\ell)new} = (A^{(\ell)h})^{-1T}\Psi_{k_\ell}^{(\ell)h}(A^{(\ell)h})^{-1}.
\end{cases}$$

\vspace{2mm}

where $A^{(\ell)h} = Var(z^{(\ell)h})$ $\forall{\ell} \in [1,L], h \in \{C, D, L_0 + 1:\}$ {and the subscript ``new" denotes the rescaled version of the parameters. %
The details are given in the Supplementary Materials. %
In the same way, the coefficients of} $\Lambda^{(0)}$ {of the discrete head are rescaled as follows: }$\Lambda^{(0) new } = \Lambda^{(0)} A^{-1T}$.

In GLLVM models, the number of coefficients of the $\Lambda^{(0)}$ matrix for binary and count data leads to a too high number of degrees of freedom. Thus, to ensure the identifiability of the model, one has to reduce the number of free coefficients. As in \cite{cagnone2014factor} the upper triangular coefficients of $\Lambda^{(0)}$ are constrained to be zero for binary and count data. This constraint is explicitly taken into account during the optimisation phase, as the optimisation program is looking for solutions for $\Lambda^{(0)}$ that are upper triangular.

\subsection{DGMM identifiability constraints}
We assume first that the latent dimension is decreasing through the layers of each head and tail \textit{i.e.} $p_h > r_1^h > ... > r_L$.
Secondly, we make the assumption that $\Lambda_{k_{\ell}}^{({\ell})hT}\Psi_{k_{\ell}}^{({\ell})-1 h}\Lambda_{k_{\ell}}^{({\ell})h}$ is diagonal with elements in decreasing order $\forall {\ell} \in [1,L]$. 
 \cite{fruehwirth2018sparse} obtained sufficient conditions for MFA identifiability,
 including the so-called \emph{Anderson-Rubin} (AR) condition, which requires that $r_{\ell} \leq \frac{r_{{\ell}-1} - 1}{2}$. Enforcing this condition would prevent from defining a MDGMM for all datasets that present less than 7 variables of each type which is far too restrictive. Then, we implement a transformation to ensure the diagonality of $\Lambda_{k_{\ell}}^{({\ell})hT}\Psi_{k_{\ell}}^{({\ell})-1 h}\Lambda_{k_{\ell}}^{({\ell})h}$ as follows:
once all parameters have been estimated by the MCEM algorithm, the following transformation is applied over $\Lambda_{k_{\ell}}^{({\ell})h}$:
\begin{itemize}
    \item Compute $B = \Lambda_{k_{\ell}}^{({\ell})hT}\Psi_{k_{\ell}}^{({\ell})-1 h}\Lambda_{k_{\ell}}^{({\ell})h}$.
    \item Decompose $B$ according to the eigendecomposition $B = P D P^{-1}$, with $D$ the matrix of the eigenvalues and $P$ the matrix of eigenvectors.
    \item Define $\Lambda_{k_{\ell}}^{({\ell})h new} = \Lambda_{k_{\ell}}^{({\ell})h} P$.
\end{itemize}

\section{Practical considerations}\label{practical_consid}

\subsection{Initialisation procedure}
\label{initialisation}
EM-based algorithms are known to be very sensitive to their initialisation values as shown for instance by \cite{biernacki2003choosing} for Gaussian Mixture models. In our case, using purely random initialization as in  \cite{cagnone2014factor} made the model diverge most of the time when the latent space was of high dimension. This can be explained by the fact that the clustering is performed in a projected continuous space of which one has no prior knowledge about. Initialising at random the latent variables $(\eta_{k_{\ell}}^{({\ell})h}, \Lambda_{k_{\ell}}^{({\ell})h}, \Psi_{k_{\ell}}^{({\ell})h}, s^{({\ell})h}, z^{({\ell})h})_{k_{\ell}, {\ell}, h}$ and the exponential family links parameters $(\lambda^{(0)}, \Lambda^{(0)})$ seems not to be a good practice. This problem gets even worse as the number of DGMM layers grows.
To stabilize our algorithm we propose the NSEP approach which combines MCA, GMM,  FA and  PLS algorithm in the M2DGMM case.
\begin{itemize}
\item For discrete head initialisation, the idea used here is to perform a Multiple Correspondence Analysis (MCA) \citep{nenadic2005computation} to determine a continuous low dimensional representation of the discrete data and use it as a first approximation of the latent variables $z^{(1)D}$.
The MCA considers all variables as categorical, thus the more the dataset actually contains this type of variables the better the initialisation should in theory be. Once this is done, a Gaussian Mixture Model is fitted in order to determine groups in the continuous space and to estimate ($\pi_{k_{\ell}}^{({\ell})}$). For each group a Factor Analysis Model (FA) is fitted to determine the parameters of the model $(\eta_{k_{\ell}}^{({\ell})}, \Lambda_{k_{\ell}}^{({\ell})}, \Psi_{k_{\ell}}^{({\ell})})$ and the latent variable of the following layer $z^{({\ell}+1)}$. 
Concerning the GLLVM parameters, logistic regressions of $y_j^{D}$ over $z^{(1)D}$ are fitted for each original variable of the discrete head: an ordered logistic regression for ordinal variables, an unordered logistic regression for binary, count and categorical variables.  
\item
For the continuous head and the common tail, the same described GMM coupled with FA procedure can be applied to determine the coefficients of the layer. The difficulty concerns the initialisation of the first tail layer with latent variable $z^{({L_0}+1)}$. Indeed, $z^{({L_0}+1)}$ has to be the same for both discrete and continuous last layers. As Factor Models are unsupervised models, one cannot enforce such a constraint on the latent variable generated from each head. To overcome this difficulty, $z^{(L_0+1)}$ has been determined by applying a PCA over the stacked variables $(z^{({L_0})C}, z^{({L_0})D})$. Then the DGMM coefficients  $(\eta_{k_{{L_0}}}^{({L_0})h}, \Lambda_{k_{{L_0}}}^{({L_0})h}, \Psi_{k_{{L_0}}}^{({L_0})}h)$ of each head have been separately determined using Partial Least Square \citep{wold2001pls} of each head last latent variable over $z^{({L_0}+1)}$.
\end{itemize}
The same ideas are used to initialize the M1DGMM. As the data going through the unique head of the M1DGMM are mixed, Factor analysis of mixed data \citep{pages2014multiple} is employed instead of MCA as it can handle mixed data.

\subsection{Model and number of clusters selection}\label{model_selection_prez}
The selection of the best MDGMM architecture is performed using the pruning methodology which is widely used in the field of supervised neural networks \citep{blalock2020state} but also for tree-based methods \citep{patil2010evaluation}.
The idea is to determine the simplest architecture that could describe the data. In order to do so, one starts with a complex architecture, and deletes the coefficients that do not carry enough information. Deleting those coefficients at some point during the training process is known as ``pre-pruning" and performing those deletions after full convergence is known as ``post-pruning". In our case, we use a pre-pruning strategy to estimate the best number of components $k_{\ell}$, the best number of factors $r_{\ell}$ and the best number of layers for the heads and tails. The reason not to use post-pruning instead of pre-pruning is that very complex architectures tend to show long running times and a higher propensity not to converge to good maxima in our simulations. 

Classical approaches {to} model specification based on information criteria, such as AIC \citep{akaike1998information} or BIC \citep{schwarz1978estimating}, need the estimation of all the possible specifications of the model. In contrast, our approach needs only one model run to determine the best architecture which is far more computationally efficient. 

In the following, we give a summary of our pruning strategy (extensive details are provided in the Supplementary Materials).
The idea is to determine the best number of components on each layer $k_{\ell}^h$ by deleting the components associated with very low probabilities $\pi_{k_{\ell}}^{({\ell})h}$ as they are the least likely to explain the data. 

The choice of the latent dimensions of each layer $r_{\ell}^h$ is performed by looking at the dimensions that carry the most important pieces of information about the previous layer. The goal is to ensure the circulation of relevant information through the layers without transmitting noise information. 
This selection is conducted differently for the GLLVM layer compared to the regular DGMM layers. 
For the GLLVM layer, we perform logistic regressions of $y^C$ over $z^{(1)C}$ and delete the dimensions that were associated with non-significant coefficients in a vast majority of paths. Concerning the regular DGMM layers, information carried by the current layer given the previous layer has been modeled using a Principal Component Analysis. We compute the contribution of each original dimension to the first principal component analysis and keep only the dimensions that present a high correlation with this first principal component, so that to drop information of secondary importance carried out through the layers. 

Finally, the choice of the total number of layers is guided by the selected $r_{\ell}$. For instance, if a dimension of two is selected for a head layer (or a dimension of one for a tail layer), then according to the identifiability constraint $p_h > r_1^h > ... > r_{\ell}^h > ... > r_L$, the following head (or tail) layers are deleted. 

Given that this procedure also selects the number of components on the tail layers, it can also be used to automatically find the optimal number of clusters in the data. The user specifies a high number of components on the clustering layer and let the automatic selection operate. The optimal number of clusters is then the number of components remaining on the clustering layer at the end of the run. This feature of the algorithm is referred to as the ``autoclus mode" of the MDGMM in the following and in the code. 

Alternatively, in case of doubt about the number of clusters in the data, the MDGMM could be used in ``multi-clustering" mode. For example, if the number of clusters in the data is assumed to be two or three, one can define a MDGMM with three components on the first tail layer and two on the second tail layer. The first layer will output a three groups clustering and the second layer a two groups clustering. The two partitions obtained can then be compared to chose the best one. This can  be done with the silhouette coefficient \citep{rousseeuw1987silhouettes} as implemented in our code. In the ``multi-clustering" mode, the same described model selection occurs. The only exception is that the number of components of the tail layers remains frozen (as {it corresponds to} the tested number of clusters in the data). 

For all clustering modes of the MDGMM, the architecture selection procedure is performed at the end of some iterations chosen by the user before launching the algorithm. Note that once the optimal specification has been determined, it is better to refit the model using the determined specification rather than keeping the former output. Indeed, changing the architecture ``on the fly" seems to disturb the quality of the final clustering. 

{Finally, in EM-based algorithms, the iteration which presents the best likelihood (the last one in general) is returned as the final output of the model. The likelihood of the model informs about how good the model is at explaining the data. However, it does not give direct information about the clustering performance of the model itself. Therefore, in the MDGMM we retain the iteration presenting the best silhouette coefficient \citep{rousseeuw1987silhouettes} among all iterations. To summarize: the likelihood criterion was used as a stopping criterion to determine the total number of iterations of the algorithm and the best silhouette score was used to select the iteration returned by the model.}

\section{Real Applications}\label{num_res}
In this section we illustrate the proposed {models} on real datasets.
First, we will present the continuous low dimensional representations of the data generated by the Discrete DGMM (DDGMM) and the {M2DGMM}. Then, the performance will be properly evaluated by comparing them to state-of-the-art mixed {data} clustering algorithms, the one-head version of the MDGMM (M1DGMM) provided with a Gaussian link function, the NSEP and the GLLVM. As some of the clustering models can deal with discrete data only (GLLVM, DDGMM) and other with mixed data (M1DGMM, MDGMM) we consider both types of data sets. The code of the introduced models is available on Github under the name MDGMM\_suite. The associated DOI is 10.5281/zenodo.4382321.

\subsection{Data description}
For the discrete data specification, we present results obtained on three datasets: the Breast cancer, the Mushrooms and the Tic Tac Toe datasets.
\begin{itemize}
\item
The Breast cancer dataset is a dataset of 286 observations and 9 discrete variables. Most of the variables are ordinal.
\item
The Tic Tac Toe dataset is composed of 9 variables corresponding to each cell of a $3\times3$ grid of tic-tac-toe. The dataset presents the grids content at the end of 958 games. Each cell can be filled with one of the player symbol (x or o), or left blanked (b) if the play has ended before all cells were filled in. Hence all the variables are categorical in contrast with the Breast cancer data.  \\
The goal is here to retrieve which game has led to victory of player 1 or of player 2 (no even games are considered here).
\item
Finally, the Mushrooms dataset is a two-class dataset with 22 attributes and 5644 observations once the missing data have been removed. The majority of the variables are categorical ones.%
\end{itemize}
For mixed datasets, we have used the Australian credit, the Heart (Statlog) and the Pima Indians diabetes Datasets.
\begin{itemize}
\item The Heart (Stalog) dataset is composed of 270 observations, five  continuous variables, three categorical variables, three binary variables and two ordinal variables. 
\item The Pima Indians Diabetes dataset presents several physiological variables (e.g. the blood pressure, the insulin rate, the age) of 768 Indian individuals. 267 individuals suffer from diabetes and the goal of classification tasks over this dataset is to distinguish the sound people from the sick ones. This dataset counts two discrete variables considered here respectively as binomial and ordinal and seven continuous variables.
\item Finally, the Australian credit (Stalog) dataset is a binary classification dataset concerning credit cards. It is composed of 690 observations, 8 discrete categorical variables and 6 continuous variables. It is a small dataset with a high dimension.
\end{itemize}
In the analysis, all the continuous variables have been centered and reduced to ensure the numeric stability of the algorithms. %
All the datasets are available in the UCI repository \citep{Dua:2019}.

\subsection{Clustering vizualisation}\label{clus_viz}
According to their multi-layer structures, the DDGMM and the {M2DGMM} perform several dimension reductions of information while the signal goes through their layers. As such, they provide low dimensional continuous representations of complex data than can be discrete, mixed or potentially highly dimensional. These representations are useful to understand how observations are clustered through the training process. They could also be reused to train other algorithms in the same spirit as for supervised Neural Networks \citep{jogin2018feature}.

\begin{figure}[H]%
\hfill
\subfigure[Training at t1]{\includegraphics[width=3.5cm]{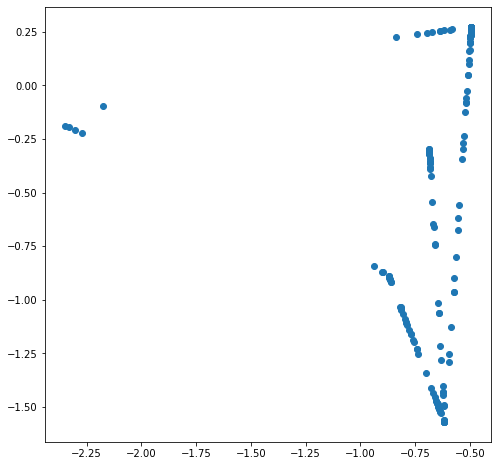}}%
\hfill
\subfigure[Training at t2]{\includegraphics[width=3.5cm]{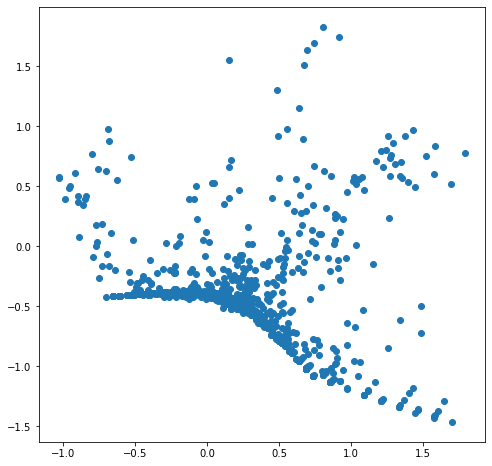}}%
\hfill
\subfigure[Training at t3]{\includegraphics[width=3.5cm]{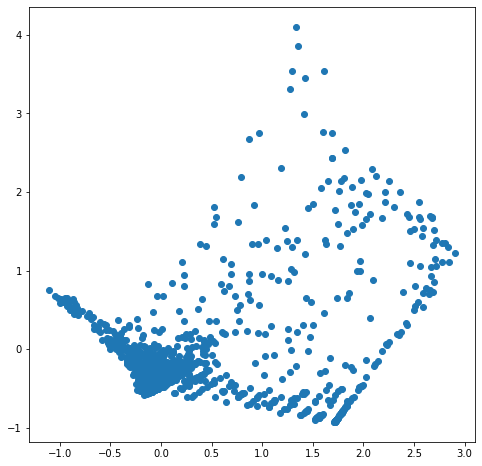}}%
\hfill
\subfigure[Training at t4]{\includegraphics[width=3.5cm]{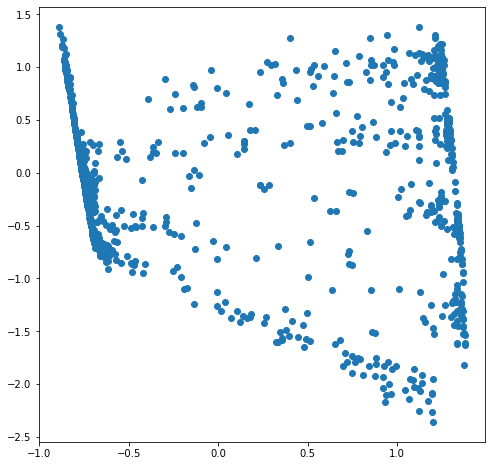}}%
\hfill
\caption{Continuous representation of the Tic Tac Toe dataset through the training of a DDGMM}%
\label{rpz_training}
\end{figure}

\begin{figure}[H]%
\subfigure[2D representation]{\includegraphics[width=8.5cm]{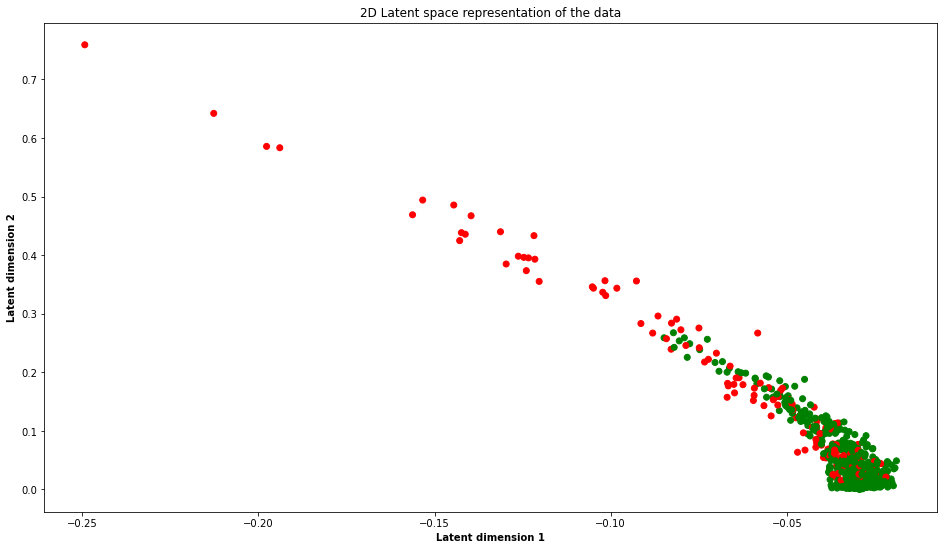}}%
\hspace{5mm}
\subfigure[3D representation]{\includegraphics[width=8.5cm]{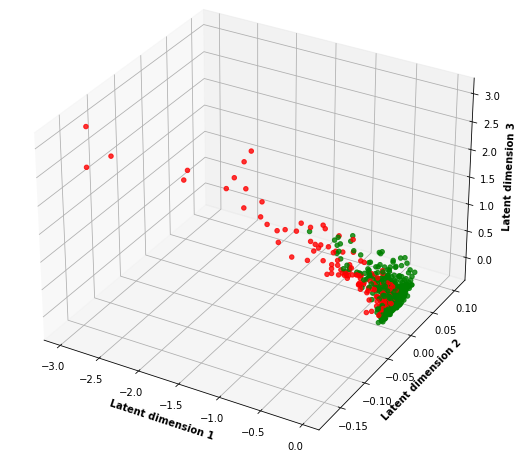}}%
\caption{Continuous representations of the Pima Diabetes dataset {provided by a M2DGMM}}%
\label{2D3D}
\end{figure}

Figure \ref{rpz_training} shows the evolution of the latent representation during the training of the clustering layer of a DDGMM for the tic tac toe dataset. Four illustrative iterations have been chosen to highlight the training process. The clustering layer has a dimension of $r_\ell = 2$ and tries to distinguish $k_\ell =2$ groups in the data.
At the beginning of the training at $t1$, it is rather difficult to differentiate two clusters in the data. However, through the next iterations, one can clearly distinguish that two sets of points are pushed away from each other by the model. Moreover in $t3$ the frontier between the two clusters can be drawn as a straight line in a two dimensional space. In $t4$ at the end of the training, the model seems to have found a simpler frontier to separate the groups as only a vertical line, i.e. a separation in a one dimensional space is needed. This highlight the information sorting process occurring through the layers in order to keep only the simplest and the more discriminating parts of the signal.
The next two figures illustrate graphical properties of the {M2DGMM}. Figure  \ref{2D3D} presents two continuous representations of the Pima Diabetes data. These are obtained during the training of a {M2DGMM} with two hidden tail layers of respectively $r_{L_0 + 1} = 3$ and $r_{L_0 + 1} = 2$ during the same iteration. Two clusters are looked for in each case ($K_{L_0+1}=K_{L_0+2}=2$) and are associated with green and red colors on the figure.
On both layers the clusters are quite well separated. The signal carried seems coherent between the two layers with a very similar structure. For the same computational cost, \textit{i.e.} one run of the model, several latent representations of the data in different dimensions can therefore be obtained.

\begin{figure}[H]%
\hfill
\subfigure[2 clusters]{\includegraphics[width=4.6cm]{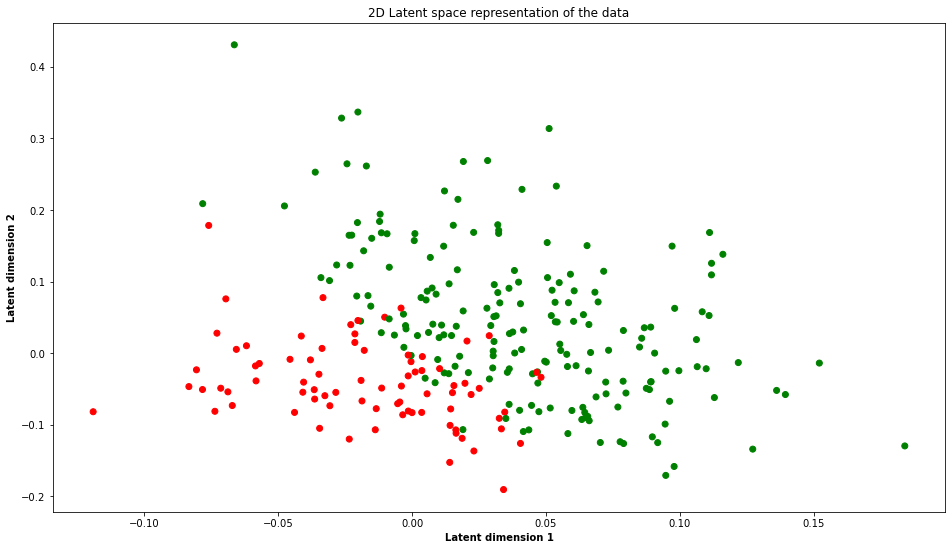}}%
\hfill
\subfigure[3 clusters]{\includegraphics[width=4.6cm]{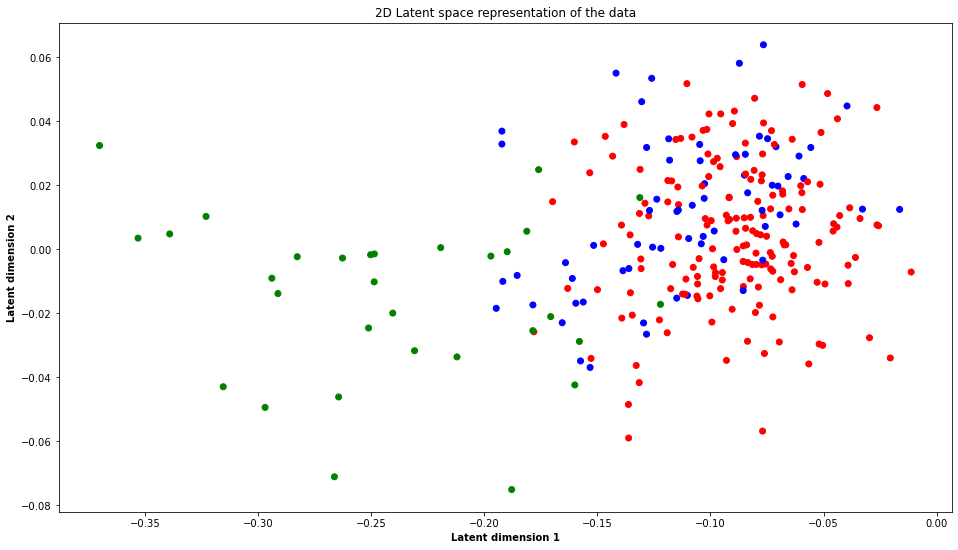}}%
\hfill
\subfigure[4 clusters]{\includegraphics[width=4.6cm]{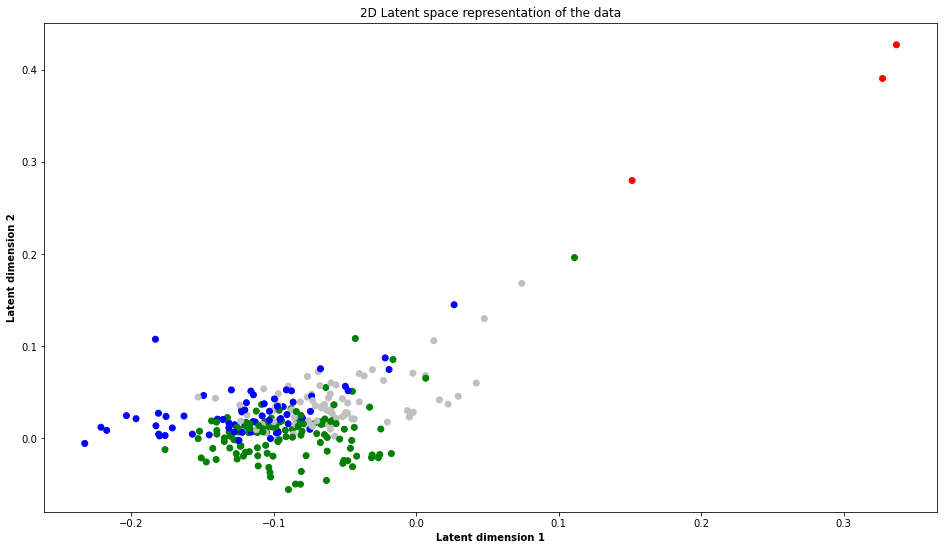}}%
\hfill
\caption{Continuous representations of the Heart dataset at the end of the training of three {M2DGMMs} with different numbers of clusters specified}%
\label{autoclus_fig}
\end{figure}

Finally, the graphical representations produced by the M2DGMM are useful tools to identify the right number of clusters in the data. Three M2DGMM have been run by setting $r_{L_0 + 1} = 2$ and with respectively $K_{L_0 + 1} = 2, K_{L_0 + 1} = 3$ and $K_{L_0 +1} = 4$. The associated latent variables are presented in Figure  \ref{autoclus_fig} with a different color for each identified cluster.

The representations with three and four clusters present points that are intertwined, with no clear distinctions between clusters. 
On the contrary, when the number of clusters searched in the data is two this separation appears distinctly. Hence, this representation advocates for a two groups distinction in the data as it is suggested by the supervised labels of the dataset (absence or presence of heart disease). The four clusters representation also shows that the three points associated with the red cluster might be outliers potentially important to study.\\
As evoked in Subsection \ref{model_selection_prez}, this visual diagnostic can be completed by using the ``autoclus mode" of the M2DGMM where the model automatically determines the best number of clusters in the data.
\subsection{Performance comparison}

\setlength\tabcolsep{0.7pt}
\captionsetup{margin=10pt,font=small,labelfont=bf}

\begin{table}[htbp]
\footnotesize
\parbox[t]{.58\linewidth}{
\centering
\begin{tabular}[t]{|p{2.2cm}|r|r|r|}
\hline
 Metrics &  \multicolumn{1}{c}{Silhouette}    &  \multicolumn{1}{c}{Micro}     &  \multicolumn{1}{c|}{Macro}     \\ \hline
   Algorithms  & \multicolumn{3}{c|}{Breast Cancer} \\ \hline
GLLVM (random init) & \ \ \ 0.215 (0.093) & \ \ \ 0.673 (0.080) & \ \ \ 0.570 (0.113) \\
GLLVM (with NSEP) & \cellcolor{LimeGreen} 0.305 (0.023) & 0.728 (0.025) & 0.671 (0.018) \\
NSEP  &  0.303 (0.000) &  0.722 (0.000) & 0.664 (0.000) \\
DDGMM    & 0.268 (0.043) & 0.696 (0.074) & 0.648 (0.048) \\
k-Modes  & 0.174 (0.000) & \cellcolor{Alizarin} 0.592 (0.000) & \cellcolor{Alizarin} 0.534 (0.000) \\
k-Prototypes & 0.293 (0.024) &  0.729  (0.014) &  0.666 (0.011) \\
Hierarchical & 0.303  (0.000) & \cellcolor{LimeGreen} 0.755 (0.000) & 0.855 (0.000) \\
SOM  & \cellcolor{Alizarin} 0.091 (0.088) & 0.668 (0.060) & 0.593 (0.011) \\
DBSCAN & 0.264 (0.000) & 0.726 (0.000) & \cellcolor{LimeGreen}0.860 (0.000) \\ \hline
            & \multicolumn{3}{c|}{Tic Tac Toe dataset} \\ \hline
GLLVM (random init) & 0.094 (0.031) & 0.591 (0.052) &  0.536 (0.100) \\
GLLVM (with NSEP)  &  0.110 (0.005) & \cellcolor{Alizarin} 0.550 (0.029) & 0.545 (0.028) \\
NSEP  &  \cellcolor{LimeGreen}0.137 (0.000) & 0.602 (0.021) & 0.597 (0.019) \\
DDGMM         & 0.118 (0.016) &  0.559 (0.028) & 0.533 (0.036) \\
k-Modes  &  0.104 (0.002) & 0.611 (0.000) & 0.586 (0.000) \\
k-Prototypes &  $\varnothing (\varnothing)$ & $\varnothing (\varnothing)$ & $\varnothing (\varnothing)$ \\
Hierarchical &  0.078 (0.000) & \cellcolor{LimeGreen}0.654 (0.000) & \cellcolor{LimeGreen}0.827 (0.000) \\
SOM  &  0.082 (0.010) & 0.650 (0.000) & 0.560 (0.000) \\
DBSCAN &   \cellcolor{Alizarin} $\varnothing (\varnothing)$ & 0.653 (0.000) & \cellcolor{Alizarin} 0.327 (0.000) \\ \hline
            & \multicolumn{3}{c|}{Mushrooms dataset} \\ \hline
GLLVM (random init) & 0.266 (0.103)  & 0.685 (0.107) & \cellcolor{Alizarin} 0.613 (0.255)\\
GLLVM (with NSEP)  &  0.351 (0.107) &  0.803 (0.102) & 0.854 (0.135) \\
NSEP  &  0.354 (0.064) & 0.811 (0.101) & 0.861 (0.074) \\
DDGMM        &  0.317 (0.078) & 0.760 (0.131) &  0.809 (0.116) \\
k-Modes  &  \cellcolor{LimeGreen}0.395 (0.000) & 0.852 (0.000) & 0.898 (0.000) \\
k-Prototypes &  0.328 (0.081) & 0.742 (0.136) & 0.818 (0.086)\\
Hierarchical &  \cellcolor{LimeGreen}0.395 (0.000) & \cellcolor{LimeGreen}0.854 (0.000) &  \cellcolor{LimeGreen}0.904 (0.000)\\
SOM  & \cellcolor{Alizarin} 0.155 (0.015) & 0.710 (0.000) & 0.814 (0.001) \\
DBSCAN &   0.294 (0.000) & \cellcolor{Alizarin} 0.624 (0.000) &  0.811 (0.000) \\ \hline
\end{tabular}
\caption{Average results and standard errors over 30 runs of the best specification for each model over three discrete datasets}\label{table1}
}
\hfill
\parbox[t]{.58\linewidth}{
\centering
\begin{tabular}[t]{|l|r|r|r|}

\hline
   Algorithms &  \multicolumn{1}{c}{Silhouette}    &  \multicolumn{1}{c}{Micro}     &  \multicolumn{1}{c|}{Macro}     \\ \hline
   Metrics      & \multicolumn{3}{c|}{Heart} \\ \hline
NSEP   & \ \ \ 0.165 (0.049) & \ \ \ 0.738 (0.068) & \ \ \ 0.739 (0.070)  \\
M1DGMM  &  0.253 (0.003) & \cellcolor{LimeGreen} 0.820 (0.012) & \cellcolor{LimeGreen} 0.820 (0.012) \\
M2DGMM   &  0.146 (0.011) & 0.710 (0.015) & 0.712 (0.014) \\
k-Modes   & 0.247 (0.000)  &  0.811 (0.000) & 0.813 (0.000) \\
k-Prototypes & \cellcolor{Alizarin} 0.044 (0.000) &  0.593 (0.000) & \cellcolor{Alizarin} 0.585 (0.000) \\
Hierarchical & \cellcolor{LimeGreen} 0.263 (0.000) & 0.811 (0.000) & 0.809  (0.000) \\
SOM  &   0.257 (0.000) & 0.795 (0.000) &  0.793 (0.000) \\
DBSCAN  & 0.177 (0.000) & \cellcolor{Alizarin} 0.556 (0.000) & 0.724 (0.000) \\ \hline
            & \multicolumn{3}{c|}{Pima} \\ \hline
NSEP   & 0.189 (0.013) & \cellcolor{LimeGreen} 0.666 (0.056) &  0.651 (0.051) \\
M1DGMM & 0.227 (0.020) & 0.633 (0.029) & 0.607 (0.029)  \\
M2DGMM   & 0.195 (0.079) & 0.647 (0.019) & 0.586 (0.068) \\
k-Modes   &  \cellcolor{Alizarin} 0.049 (0.033) & \cellcolor{Alizarin} 0.581 (0.000) &  \cellcolor{Alizarin} 0.482 (0.000) \\
k-Prototypes & $\varnothing$ ($\varnothing$) & $\varnothing$ ($\varnothing$) & $\varnothing$ ($\varnothing$) \\
Hierarchical &   \cellcolor{LimeGreen} 0.391 (0.000)  &  0.656 (0.000) & \cellcolor{LimeGreen} 0.826 (0.000)  \\
SOM  & 0.232 (0.000) & 0.644 (0.000) & 0.610 (0.003) \\
DBSCAN  & \cellcolor{LimeGreen} 0.391 (0.000) &  0.654 (0.000) & \cellcolor{LimeGreen} 0.826 (0.000)  \\ \hline
                        & \multicolumn{3}{c|}{Australian Credit} \\ \hline
NSEP   & 0.165 (0.034) & 0.754 (0.098) &  0.753 (0.110) \\
M1DGMM  &  0.170 (0.032) &	0.707 (0.112) & 0.806 (0.036) \\
M2DGMM   & 0.224 (0.080) &  0.575 (0.040) &  0.680 (0.104) \\
k-Modes   &  0.222 (0.007) & 0.785 (0.008) & 0.784  (0.007) \\
k-Prototypes &  0.163 (0.000) & \cellcolor{Alizarin} 0.562 (0.000)  & 0.780 (0.000)\\
Hierarchical &  \cellcolor{LimeGreen}0.399 (0.000) & \cellcolor{LimeGreen}0.849 (0.000) & \cellcolor{LimeGreen}0.847 (0.000)\\
SOM  &  \cellcolor{Alizarin} 0.127 (0.096) & 0.649 (0.001) &\cellcolor{Alizarin} 0.676 (0.002) \\
DBSCAN  &  0.201 (0.000)  &  0.570  (0.000) &  0.740 (0.000)  \\ \hline
\end{tabular}
\caption{Average results and standard errors over 30 runs of the best specification for each model over three mixed datasets}\label{table2}
}%
\end{table}%
In order to benchmark the performance of the proposed strategy, we consider alternative algorithms coming from {each family of approaches identified by \citet{ahmad2019survey}}, namely k-modes, k-Prototypes, Hierarchical Clustering, Self-Organising Maps (SOM), and DBSCAN \citep{ester1996density}.

For each dataset, we have set the number of unsupervised clusters to the ``ground truth" classification number. In order to present a fair report, several specifications of the benchmark models have been run. For each specification, the models have been launched 30 times. The reported results correspond to the best specification of each benchmark model with respect to each metric on average over the 30 runs.
The set of specifications evaluated for each benchmark model is given in the Supplementary Materials. %
Concerning our models, the architectures were automatically selected and then fitted 30 times on each dataset.\\
Here we use one unsupervised metric and two supervised metrics to assess the clustering quality: the silhouette coefficient, the micro precision and the macro precision. The silhouette coefficient measures how close on average a point is from the points of the same group with respect to the points of the other groups. The Euclidian distance cannot be used here due to the mixed feature space and hence the Gower distance \citep{gower1971general} is used instead. The silhouette coefficient ranges between 1 (perfect clustering) and -1 (meaningless clustering). The micro precision corresponds to the overall accuracy, \emph{i.e.} the proportion of correctly classified instances. The macro precision computes the proportion of correctly classified instances per class and then returns a non-weighted mean of those proportions. These two quantities tend to differ when the data are not balanced. The formal expressions of the metrics are given in the Supplementary Materials. Note that we cannot use AIC or BIC criteria here since their values are not available for all methods.  %

Tables \ref{table1}-\ref{table2} present the best average results obtained by the algorithms and the associated standard error over the 30 runs in parenthesis. The best algorithm for a given dataset and metric is associated with a green cell and the worst with a red cell. An empty set symbol means that the metric was not defined for this algorithm on that dataset. For the special case of the k-prototypes algorithm, the empty set symbol means that the dataset contained only one type of discrete data which is a situation that the algorithm is not designed for.

\subsubsection{Results on discrete data}
The new initialisation (NSEP) enables the GLLVM to achieve better performances on the Mushrooms dataset and on the Breast dataset where the GLLVM attains the best silhouette score. It also stabilizes the GLLVM as the standard errors obtained are divided by at least a factor two for all metrics of the Breast Cancer and of the Tic Tac Toe datasets. \\%
The NSEP in itself gives good results for all metrics and is often among the best two performing models. Finally, over the Tic Tac Toe dataset the DDGMM performs slightly better than the GLLVM, but less on the two other datasets.
 \\
{Hence, compared to the other methods, the models introduced in this work represent solid baseline models. On the contrary, some alternative methods appear to fit some datasets well and poorly other ones. This is the case for instance of DBSCAN which performs well on the Breast cancer dataset, but much less on the Mushrooms and the Tic Tac Toe datasets (the algorithm could find only one group in the Tic Tac Toe data which explains that the silhouette score is not defined). Another example is k-Modes which obtains substantial results on the Mushrooms dataset but under-average results for the two other datasets. Finally, among all methods, the hierarchical clustering is the algorithm that performs best on a majority of metrics and datasets.}

\subsubsection{Results on mixed data}
As clear from results in Table \ref{table2}, the NSEP seems again to be a good starting point for both algorithms and certainly also explains the fact that the M1DGMM reaches the best micro and macro scores on the Heart dataset.\\
The M1DGMM achieves better average results than the M2DGMM except for the silhouette score on the Australian Credit dataset and the micro precision on the Pima dataset. The two specifications tend to often present opposite patterns in terms of standard errors: when the M2DGMM results are stable the M1DGMM results tend to be more volatile and vice versa. Hence, depending on whether one wants to minimize the bias or the variance of the estimation, the two specifications seem complementary and could be used in turn to conduct clustering on a large diversity of datasets.\\
As in the discrete data results, the models introduced and especially the M1DGMM, give satisfactory performance on all datasets on average, whereas other models such as SOM, DBSCAN or k-modes perform well on some datasets only. Similarly, the hierarchical clustering method seems to provide the best results on a large set of metrics and datasets.

\section{Conclusion}\label{conclusion}
This work aimed to provide a  reliable and flexible model for clustering mixed data by borrowing ingredients from the GLLVM and the DGMM recent approaches.
Several sub-models have been introduced and could be used on their own:
\begin{itemize}
\item
 a new initialisation procedure called NSEP for GLLVM-based models,
 \item
 a Discrete DGMM (DDGMM) for discrete data,
 \item
 a one-head (M1DGMM) and a two-heads (M2DGMM) DGMM for mixed data.
 \end{itemize}
This suite of models handles the usual clustering issues concerning architecture selection and the choice of the number of clusters in the data in an automated manner.\\
From the experiments carried out on real data, the MDGMM performances are in line with the other state-of-the-art models. It can be regarded as a baseline model over a general class of data. Its use of nested Mixtures of Factor Analyzers enables it to capture a very wide range of distributions and patterns.\\
Despite of its complexity, the MDGMM remains interpretable. From a practical viewpoint, the structure of the latent space can be observed through the model training with the help of the graphical utilities presented in section \ref{clus_viz}. Thus, they allow the user to perform visual diagnostics of the clustering process. From a theoretical standpoint, the parameters of the model remain interpretable as the link between parameters and clustering results is proper thanks to the identifiability of the model. The set of identifiability constraints presented here could seem quite restrictive. However, it forces the model to stay in a quite well delimited parameter space and to avoid for instance a too significant explosion of the norm of the parameters values.
The implementation of these constraints can nevertheless be improved by considering a Bayesian re-writing of our model on Variational principles. Indeed, it should make identification requirements easier to meet, as one can keep only the posterior draws that meet the identifiability requirements.  \cite{niku2019efficient} have rewritten the GLLVM model in a variational fashion and exhibit high running time and accuracy gains. Following their path, one could adapt the MDGMM to the variational framework.\\
Finally considering the training process, the choice of an EM-based algorithm was motivated by its extensive use in the Gaussian Mixture Model literature. The EM-related algorithms are however very sensitive to the initialisation, which was in our case particularly tricky given the size of the parameter space. Combining Multiple Correspondence Analysis with Gaussian Mixture Models, Factor Analysis  and Partial Least Squares into NSEP has however enabled us to significantly stabilize the estimation process. Yet, new initialisation and training processes could be designed to help the model to better rationalize latent structures in the data within its very highly dimensional space.


\section*{Acknowledgments}
Thanks to the LIA LYSM (agreement between AMU, CNRS, ECM and INdAM) for having funded a mission to Bologna.
Thanks also to Nicolas Chopin and Samuel Soubeyrand for their helpful advices.


\bibliographystyle{Chicago}

\bibliography{sources}

  \section{Supplementary Material}
\label{supplement}
\subsection{Expression of the expected Log-Likelilhood}
The expected log-likelihood  can be expressed as:
\begin{align*}
&\mathbb{E}_{z^{C}, z^{D}, z^{({L_0} + 1:)},  s^C, s^D, s^{({L_0}+1:)} | y^C, y^D,  \hat{\Theta}_C, \hat{\Theta}_D, \hat{\Theta}_{{L_0}+1:}}[\log L(y^C, y^D, z^{C}, z^{D}, z^{({L_0} + 1:)},  s^C, s^D, s^{({L_0}+1:)} | \Theta_C, \Theta_D, \Theta_{{L_0} + 1:})]
\\
&= \mathbb{E}_{z^{(1)D}, s^D, s^{({L_0}+1:)D} | y^D, \hat{\Theta}_D, \hat{\Theta}_{{L_0}+1:}}[\log L(y^D | z^{(1)D}, s^D, s^{({L_0}+1:)}, \Theta_D, \Theta_{{L_0} + 1:})] \\
&+ \mathbb{E}_{z^{(1)C}, s^C, s^{({L_0}+1:)C} | y^C, \hat{\Theta}_C, \hat{\Theta}_{{L_0}+1:}}[\log L(y^C | z^{(1)C}, s^C, s^{({L_0}+1:)}, \Theta_C, \Theta_{{L_0} + 1:})] \\
&+ \sum_{h \in \{C, D\}} \sum_{l=1}^{{L_0}} \mathbb{E}_{z^{(l)h}, z^{(l+1)h}, s^h, s^{({L_0}+1:)} | y^h,  \hat{\Theta}_h, \hat{\Theta}_{{L_0}+1:}}[\log L(z^{(l)h} | z^{(l + 1)h}, s^h, s^{({L_0}+1:)}, \Theta_h, \Theta_{({L_0}+1:)})] \\
&+ \sum_{l = {L_0} + 1}^{L-1} \mathbb{E}_{z^{(l)}, z^{(l+1)}, s^C, s^D, s^{({L_0}+1:)} | y^C, y^D,  \hat{\Theta}_C, \hat{\Theta}_D, \hat{\Theta}_{{L_0}+1:}}[\log L(z^{(l)} | z^{(l + 1)}, s^C, s^D, s^{({L_0}+1:)}, \Theta_C, \Theta_D, \Theta_{{L_0}+1:}] \\
&+ \mathbb{E}_{z^{(L)} | y^C, y^D, \hat{\Theta}_C, \hat{\Theta}_D, \hat{\Theta}_{{L_0}+1:}}[\log L(z^{(L)} | \Theta_C, \Theta_D, \Theta_{{L_0} + 1:}))] \\
&+ \mathbb{E}_{s^C, s^D, s^{({L_0} + 1:)}|y^C, y^D, \hat{\Theta}_C, \hat{\Theta}_D, \hat{\Theta}_{{L_0}+1:}}[\log L(s^C, s^D, s^{({L_0} + 1:)} | \Theta_C, \Theta_D, \Theta_{{L_0} + 1:})],
\numberthis \label{exp_ell}
\end{align*}
%
with a slight abuse of notation in the double sum as we have set  $z^{({L_0} + 1)} = z^{({L_0} + 1)C} = z^{({L_0} + 1)D}$. $\hat{\Theta}_h$ are the provisional estimate of $\Theta_h$ through the iterations of the algorithm.
\subsection{GLLVM embedding layer mathematical derivations}
\subsubsection{E step for the GLLVM embedding layer}

We consider the conditional density
\begin{align}
        f(z^{(1)D} | y^D, \hat{\Theta}_D, \hat{\Theta}_{{L_0} + 1:})  = \sum_{s'}f(z^{(1)D} | y^D, s', \hat{\Theta}_D, \hat{\Theta}_{{L_0} + 1:})f(s^{(1D:L)} = s'| y^D, \hat{\Theta}_D, \hat{\Theta}_{{L_0} + 1:}).
        \label{p(zs_y)}
\end{align}
%
The Bayes rule for the first term gives :
\begin{equation}
    f(z^{(1)D} | y^D, s', \hat{\Theta}_D, \hat{\Theta}_{{L_0} + 1:}) = \frac{f(z^{(1)D}| s', \hat{\Theta}_D, \hat{\Theta}_{{L_0} + 1:})f(y^D | z^{(1)D}, \hat{\Theta}_D, \hat{\Theta}_{{L_0} + 1:})}{f(y^D | s', \hat{\Theta}_D, \hat{\Theta}_{{L_0} + 1:})}
    \label{p(z_ys)},
\end{equation}
and we have
$$(z^{(1)D}| s', \hat{\Theta}_D, \hat{\Theta}_{{L_0} + 1:}) \sim N(\mu_{s'}^{(1D:L)}, \Sigma_{s'}^{(1D:L)}),$$
where the mean and covariance parameters $(\mu_{s'}^{(1D:L)}, \Sigma_{s'}^{(1D:L)})$  are detailed in Section \ref{E-STEP-DGMM}.\\
Moreover,  $f(y^D | z^{(1)D}, \hat{\Theta}_D, \hat{\Theta}_{{L_0} + 1:})$ belongs to an exponential family.  
Finally, $f(y^D | s', \hat{\Theta}_D, \hat{\Theta}_{{L_0} + 1:})$ has to be numerically approximated. This is here performed by Monte Carlo estimation by simulating $M^{(1)}$ copies of $z^{(1)D}$ as follows
\begin{align*}
    f(y^D|s', \hat{\Theta}_D, \hat{\Theta}_{{L_0} + 1:}) &=  \int_{z^{(1)D}} f(y^D | z^{(1)D}, \hat{\Theta}_D, \hat{\Theta}_{{L_0} + 1:})f(z^{(1)D} | s', \hat{\Theta}_D, \hat{\Theta}_{{L_0} + 1:})dz^{(1)D} \\
    &\approx \sum_{m=1}^{M^{(1)}}  f(y^D | z_m^{(1)D}, \hat{\Theta}_D, \hat{\Theta}_{{L_0} + 1:}, \hat{\Theta})f(z_m^{(1)D} | s', \hat{\Theta}_D, \hat{\Theta}_{{L_0} + 1:}).
\end{align*}
The second term of (\ref{p(zs_y)}) can be written as a posterior density:
\begin{equation}
     f(s^{(1D:L)} = s' | y^D, \hat{\Theta}_D, \hat{\Theta}_{{L_0} + 1:}) =  \frac{f(s^{(1D:L)} = s' | \hat{\Theta}_D, \hat{\Theta}_{{L_0} + 1:})f(y^D| s', \hat{\Theta}_D, \hat{\Theta}_{{L_0} + 1:})}{\sum_{s''} f(s^{(1D:L)} = s''| \hat{\Theta}_D, \hat{\Theta}_{{L_0} + 1:})f(y^D| s^{(1D:L)} = s'', \hat{\Theta}_D, \hat{\Theta}_{{L_0} + 1:})}
     \label{ps_y},
\end{equation}
\vspace{2mm}%
and we have $(s^{(1D:L)} | \hat{\Theta}_D, \hat{\Theta}_{{L_0} + 1:}) \sim M(\pi_{s}^{(1D:L)})$ a multinomial distribution with parameters $\pi_{s}^{(1D:L)}$ which is the probability of a full path through the network starting from the discrete head.  The density $f(y^D| s', \hat{\Theta}_D, \hat{\Theta}_{{L_0} + 1:})$  is once again approximated by Monte Carlo.

\subsubsection{M step for the GLLVM embedding layer}
To maximize $\mathbb{E}_{z^{(1)D}| y^D, \hat{\Theta}_D, \hat{\Theta}_{L_0 + 1:}}[\log L(y^D| z^{(1)D}, \Theta_D, \hat{\Theta}_{L_0 + 1:})]$, we use optimisation methods. All methods belong to the Python scipy.optimize package \citep{2020SciPy-NMeth}. 
For binary, count and categorical variables, the optimisation program is unconstrained and the BFGS \citep{fletcher2013practical} algorithm is used. Concerning ordinal variables, the optimisation program is constrained as the intercept coefficients have to be ordered. The method used is a trust-region algorithm \citep{conn2000trust}. 
All the gradients are computed by automatic differentiation using the autograd package \citep{maclaurin2015autograd}, which significantly speeds up the optimization process compared to hand-coded gradients. 

\subsection{DGMM layers mathematical derivations}
\subsubsection{E step for the DGMM layers}
\label{E-STEP-DGMM}
Recall that we have:
\begin{align}
    f(z^{(\ell)}, z^{(\ell+1)}, s | y,  \hat{\Theta})
    &= f(z^{(\ell)}, s | y,  \hat{\Theta}) f(z^{(\ell+1)} | z^{(\ell)}, s, y, \hat{\Theta}) \nonumber \\
    &= f(z^{(\ell)} | y,  s, \hat{\Theta}) f(s | y, \hat{\Theta}) f(z^{(\ell+1)} | z^{(\ell)}, s, \hat{\Theta}).
    \label{E_distrib_l}
\end{align}
%
The first term can be rewritten and approximated as follows:
\begin{align}
    f(z^{(\ell)} | y,  s, \hat{\Theta}) &=  \int_{z^{(\ell-1)}}  f(z^{(\ell)} | z^{(\ell-1)},  s, \hat{\Theta}) f(z^{(\ell-1)} | y,  s, \hat{\Theta}) dz^{(\ell-1)} \nonumber \\
    &\approx \sum_{m = 1}^{M^{(\ell-1)}} f(z^{(\ell)} | z_m^{(\ell-1)},  s, \hat{\Theta}) f(z_m^{(\ell-1)} | y,  s, \hat{\Theta}).
    \label{rec_comput}
\end{align}
This expression is hence calculable in a recurrent manner $\forall{{\ell}} \in [2,{L_0}]$, starting with $f(z^{(1)} | y, s', \hat{\Theta})$
given by (\ref{p(z_ys)}).
The second term of (\ref{E_distrib_l}) can be expressed as  in (\ref{ps_y}), 
and the last term  
is given by the Bayes rule:
\begin{align}
    f(z^{({\ell}+1)} | z^{({\ell})}, s, \hat{\Theta}) &= \frac{f(z^{({\ell})} | z^{({\ell}+1)}, s, \hat{\Theta})f(z^{({\ell}+1)} | s, \hat{\Theta})}{f(z^{({\ell})} | s, \hat{\Theta})}.  
    \label{pz2_z1s}
\end{align}
Clearly, the denominator does not depend on $z^{({\ell}+1)}$ and is hence considered as a normalisation constant. Besides, we have that $f(z^{({\ell})} | z^{({\ell}+1)}, s, \hat{\Theta}) = N(\eta_{k_{\ell}}^{({\ell})} + \Lambda_{k_{\ell}}^{({\ell})} z^{({\ell}+1)}, \Psi_{k_{\ell}}^{({\ell})})$. 
Finally, by construction of the DGMM, we have
\begin{align}
    f(z^{({\ell}+1)} | s, \hat{\Theta}) = f(z^{({\ell}+1)} | s^{(l+1:L)}, \hat{\Theta}) 
    = N(\mu_{s^{(:k_{{\ell}+1}:)}}^{({\ell}+1)}, \Sigma_{s^{(:k_{{\ell}+1}:)}}^{({\ell}+1)}).
    \label{gauss_distrib}
\end{align}
%
It follows that (\ref{pz2_z1s}) is also a Gaussian distribution of parameters $(\rho_{k_{{\ell}+1}}^{({\ell}+1)}, \xi_{k_{{\ell}+1}}^{({\ell}+1)})$.

The formulas of the Gaussian parameters are obtain as follows:
the DGMM can be written at each layer as a regular Gaussian Mixture with a number of components equal to the number of paths starting from that layer. The Gaussian mean and covariance matrix of each path starting from the $k_{\ell}$ component of layer $\ell$ can be computed in the following way:

\begin{align*}
\mu_{\tilde{s}^{(k_{{\ell}}:)}}^{({\ell})} &= \eta_{k_{{\ell}}}^{({\ell}+1)} +  \sum_{j = {\ell} + 1 }^{L}\big(\prod_{m={\ell}}^{j-1}\Lambda_{k'_m}^{(m)}\big)\eta_{k'_j}^{(j)},
\end{align*}
and
\begin{align*}
\Sigma_{\tilde{s}^{(k_{{\ell}}:)}}^{({\ell})} &= \Psi_{k_{{\ell}}}^{({\ell})} +  \sum_{j= {\ell} + 1}^{L}\big(\prod_{m={\ell}}^{j-1}\Lambda_{k'_m}^{(m)}\big)(\Psi_{k'_j}^{(j)} + \Lambda_{k'_{\ell}}^{(j)}\Lambda_{k'_j}^{(j)T})\big(\prod_{m={\ell}}^{j-1}\Lambda_{k'_m}^{(m)}\big)^T.
\end{align*}

In addition, we have that the random variable $(z^{({\ell}+1)} | z^{({\ell})}, \tilde{s}, \hat{\Theta})$ also follows a multivariate Gaussian distribution with mean and covariance parameters $(\rho_{k_{\ell+1}}^{({\ell}+1)}, \xi_{k_{{\ell}+1}}^{({\ell}+1)})$:

\[
    \rho_{k_{{\ell}+1}}^{({\ell}+1)} = \xi_{k_{{\ell}+1}}^{({\ell}+1)}\Big(\Lambda_{k_{{\ell}+1}}^{({\ell}+1)T}(\Psi_{k_{{\ell}+1}}^{({\ell}+1)})^{-1}(z^{({\ell})} - \eta_{k_{{\ell}+1}}^{({\ell}+1)}) + \Sigma_{\tilde{s}^{(:k_{{\ell}+1}:)}}^{({\ell}+1)}\mu_{\tilde{s}^{(:k_{{\ell}+1}:)}}^{({\ell}+1)} \Big),
\]

and

\[
    \xi_{k_{{\ell}+1}}^{({\ell}+1)} = \Big(\Sigma_{\tilde{s}^{(:k_{{\ell}+1}:)}}^{({\ell}+1)} + \Lambda_{k_{{\ell}+1}}^{({\ell}+1)T}(\Psi_{k_{{\ell}+1}}^{({\ell}+1)})^{-1}\Lambda_{k_{{\ell}+1}}^{({\ell}+1)}\Big)^{-1}.
\]

\subsubsection{M Step for the DGMM layers}

We now turn on to the log-likelihood expression and give the estimators of the $\ell$-th DGMM layer parameters $\forall{\ell \in [1,L_h]}$, $\forall{h} \in \{C, D\}$. In this section the $h$ superscripts are omitted for simplicity of notation.
\begin{multline*}
\log L(z_i^{({\ell})} | z_i^{({\ell}+1)}, s_i, \Theta) = \\
- \frac{1}{2} \left[ \log(2\pi) +  \log\det(\Psi_{k_{{\ell}}}^{({\ell})}) + \Big{(}z_i^{({\ell})} - (\eta_{k_{{\ell}}}^{({\ell})} +  \Lambda_{k_{{\ell}}}^{({\ell})}z_i^{({\ell}+1)})\Big{)}^T\Psi_{k_{{\ell}}}^{({\ell})-1}\Big{(}z_i^{({\ell})} - (\eta_{k_{{\ell}}}^{({\ell})} +  \Lambda_{k_{{\ell}}}^{({\ell})}z_i^{({\ell}+1)})\Big{)}\right].
\end{multline*}
The derivatives of this quantity with respect to $\eta_{k_{{\ell}}}^{({\ell})}, \Lambda_{k_{{\ell}}}^{({\ell})}, \Psi_{k_{{\ell}}}^{({\ell})}$ are given by

$$
\begin{cases}
\frac{\partial{\log L(z_i^{({\ell})} | z_i^{({\ell}+1)}, s_i, \Theta)}}{\partial{\eta_{k_{{\ell}}}^{({\ell})}}} = \Psi^{({\ell})-1}_{k_{{\ell}}}\Big{(}z_i^{({\ell})} - (\eta_{k_{{\ell}}}^{({\ell})} +  \Lambda_{k_{{\ell}}}^{({\ell})}z_i^{({\ell}+1)})\Big{)}\\

\frac{\partial{\log L(z_i^{({\ell})} | z_i^{({\ell}+1)}, s_i, \Theta)}}{\partial{\Lambda_{k_{{\ell}}}^{({\ell})}}} = \Psi^{({\ell})-1}_{k_{{\ell}}}\Big{(}z_i^{({\ell})} - (\eta_{k_{{\ell}}}^{({\ell})} +  \Lambda_{k_{{\ell}}}^{({\ell})}z_i^{({\ell}+1)})\Big{)}z_i^{({\ell}+1)T}\\

\frac{\partial{\log L(z_i^{({\ell})} | z_i^{({\ell}+1)}, s_i, \Theta)}}{\partial{\Psi_{k_{{\ell}}}^{({\ell})}}} =
-\frac{1}{2}\Psi^{({\ell})-1}_{k_{{\ell}}}\left[I_{r_1}- \Big{(}z_i^{({\ell})} - (\eta_{k_{{\ell}}}^{({\ell})} +  \Lambda_{k_{{\ell}}}^{({\ell})}z_i^{({\ell}+1)})\Big{)}\Big{(}z_i^{({\ell})} - (\eta_{k_{{\ell}}}^{({\ell})} +  \Lambda_{k_{{\ell}}}^{({\ell})}z_i^{({\ell}+1)})\Big{)}^T\Psi^{({\ell})-1}_{k_{{\ell}}}\right].
\end{cases}
$$
\vspace{5mm}%
Taking the expectation of the derivative with respect to $\eta_{k_{{\ell}}}^{({\ell})}$ and equalizing it to zero, it follows that:
\begin{align*}
    &\mathbb{E}_{z^{({\ell})}, z^{({\ell}+1)}, s | y,  \hat{\Theta}}\left[\frac{\partial{\log L(z^{({\ell})} | z^{({\ell}+1)}, s, \Theta)}}{\partial{\eta_{k_{{\ell}}}^{({\ell})}}}\right] = 0 \\
    &\iff \Psi^{({\ell})-1}_{k_{{\ell}}}\sum_{i=1}^n \mathbb{E}_{z_i^{({\ell})}, z_i^{({\ell}+1)}, s_i| y_i,  \hat{\Theta}}\left[z_i^{({\ell})} - (\eta_{k_{{\ell}}}^{({\ell})} + \Lambda_{k_{{\ell}}}^{({\ell})}z_i^{({\ell}+1)}) \right] = 0 \\
    &\iff \sum_{i=1}^n \mathbb{E}_{z_i^{({\ell})}, z_i^{({\ell}+1)}, s_i | y_i,  \hat{\Theta}}\left[ z_i^{({\ell})} - (\eta_{k_{{\ell}}}^{({\ell})} + \Lambda_{k_{{\ell}}}^{({\ell})}z_i^{({\ell}+1)}) \right] = 0,   \text{ since } \Psi^{({\ell})}_{k_{{\ell}}} \text{ is positive semi-definite.}\\
    &\iff \sum_{i=1}^{n} \sum_{\tilde{s}_i^{(:k_{\ell}:)}} f(s_i^{(:k_{\ell}:)} = \tilde{s}_i^{(:k_{\ell}:)}| y_i, \hat{\Theta}) \left[  \mathbb{E}_{z_i^{({\ell})}| \tilde{s}_i^{(:k_{\ell}:)}, y_i,  \hat{\Theta}}[z_i^{({\ell})} ] - \eta_{k_{{\ell}}}^{({\ell})}  - \Lambda_{k_{{\ell}}}^{({\ell})}\mathbb{E}_{z_i^{({\ell}+1)}| \tilde{s}_i^{(:k_{\ell}:)}, y_i,  \hat{\Theta}}[ z_i^{({\ell}+1)}]\right] = 0. 
\end{align*}
Therefore, the estimator of $\eta_{k_{{\ell}}}^{({\ell})}$ is given by
\[
    \hat{\eta}_{k_{{\ell}}}^{({\ell})} = \frac{\sum_{i=1}^{n} \sum_{\tilde{s}_i^{(:k_{\ell}:)}} f(s_i^{(:k_{\ell}:)} = \tilde{s}_i^{(:k_{\ell}:)}|y, \hat{\Theta})\left[
    E[z_i^{({\ell})} | s_i^{(:k_{\ell}:)} = \tilde{s}_i^{(:k_{\ell}:)}, y_i,  \hat{\Theta}] - \Lambda_{k_{{\ell}}}^{({\ell})}E[z_i^{({\ell}+1)}| \tilde{s}_i^{(:k_{\ell}:)},  y_i,  \hat{\Theta}]\right]}{\sum_{i=1}^{n} \sum_{\tilde{s}_i^{(:k_{\ell}:)}} f(s_i^{(:k_{\ell}:)} = \tilde{s}_i^{(:k_{\ell}:)}|y_i, \hat{\Theta})},
\]
with
\begin{align*}
    E[z_i^{({\ell}+1)} | s_i^{(:k_{\ell}:)} = \tilde{s}_i^{(:k_{\ell}:)},  y_i,  \hat{\Theta}] &= \int_{z_i^{({\ell})}} f(z_i^{({\ell})} | \tilde{s}_i^{(:k_{\ell}:)}, y_i, \hat{\Theta}) \int_{z_i^{({\ell}+1)}} f(z_i^{({\ell}+1)} | z_i^{({\ell})}, \tilde{s}_i^{(:k_{\ell}:)}, \hat{\Theta})z_i^{({\ell}+1)} dz_i^{({\ell}+1)} dz_i^{({\ell})}\\
    &\approx \sum_{m_{{\ell}}=1}^{M^{({\ell})}} f(z_{i, m_{{\ell}}}^{({\ell})} | \tilde{s}_i^{(:k_{\ell}:)}, y_i, \hat{\Theta}) \sum_{m_{{\ell}+1}=1}^{M^{({\ell}+1)}} z_{i, m_{{\ell}+1}}^{({\ell}+1)},
\end{align*}
where $z_{i, m_{{\ell}+1}}^{({\ell}+1)}$ has been drawn from $f(z_{i,m_{{\ell}+1}}^{({\ell}+1)} | z_{i,m_{\ell}}^{({\ell})}, s)$.
\vspace{10mm}%
Using the same reasoning for $\Lambda_{k_{{\ell}}}^{({\ell})}$ we obtain
\begin{align*}
    &\mathbb{E}_{z^{({\ell})}, z^{({\ell}+1)}, s | y,  \hat{\Theta}}\left[\frac{\partial{\log L(z^{({\ell})} | z^{({\ell}+1)}, s, \Theta)}}{\partial{\Lambda_{k_{{\ell}}}^{({\ell})}}}\right] = 0 \\
    &\iff \Psi^{({\ell})-1}_{k_{{\ell}}}\sum_{i=1}^n\left[\mathbb{E}_{z_i^{({\ell})}, z_i^{({\ell}+1)}, s_{i} | y_i,  \hat{\Theta}}[(z_i^{({\ell})} - (\eta_{k_{{\ell}}}^{({\ell})} + \Lambda_{k_{{\ell}}}^{({\ell})}z_i^{({\ell}+1)}))z_i^{({\ell}+1)T}] \right]= 0\\
    &\iff \sum_{i=1}^n \sum_{\tilde{s}_i^{(:k_{\ell}:)}} f(s_i^{(:k_{\ell}:)} = \tilde{s}_i^{(:k_{\ell}:)}| y_i, \hat{\Theta})\Big[
    \mathbb{E}_{z_i^{({\ell})}, z_i^{({\ell}+1)} | \tilde{s}_i^{(:k_{\ell}:)}, y_i,  \hat{\Theta}}
    \\
    &  ~~~~~~~~~~~~~~~~~~~~ [(z_i^{({\ell})} - \eta_{k_{l}}^{({\ell})})z_i^{({\ell}+1)T}]
    - \Lambda_{k_{{\ell}}}^{({\ell})}\mathbb{E}_{z_i^{({\ell}+1)} | \tilde{s}_i^{(:k_{\ell}:)}, y_i,  \hat{\Theta}}[z_i^{({\ell}+1)}z_i^{(l+{\ell})T}] \Big] \\
    & \ \ \ \ \ \ \ \ \ = 0. 
\end{align*}
Hence the estimator of $\Lambda_{k_{{\ell}}}^{({\ell})}$ is given by
\[
    \hat{\Lambda}_{k_{{\ell}}}^{({\ell})} =
    \frac{\sum_{i=1}^n \sum_{\tilde{s}_i^{(:k_{\ell}:)}} f(s_i^{(:k_{\ell}:)} = \tilde{s}_i^{(:k_{\ell}:)}| y_i, \hat{\Theta}) \Big[E[(z_i^{({\ell})}- \hat{\eta}_{k_{{\ell}}}^{({\ell})})z_i^{({\ell}+1)T} | \tilde{s}_i^{(:k_{\ell}:)}, y_i,  \hat{\Theta}]\Big]}{\sum_{i=1}^n \sum_{\tilde{s}_i^{(:k_{\ell}:)}} f(s_i^{(:k_{\ell}:)} = \tilde{s}_i^{(:k_{\ell}:)}| y_i, \hat{\Theta})}E[z_i^{({\ell}+1)}z_i^{({\ell}+1)T} | \tilde{s}_i^{(:k_{\ell}:)}, y_i, \hat{\Theta}]^{-1},
\]
with
\begin{small}
\begin{align*}
    E[(z_i^{({\ell})}- \hat{\eta}_{k_{{\ell}}}^{({\ell})})z_i^{({\ell}+1)T} | \tilde{s}_i^{(:k_{\ell}:)}, y_i,  \hat{\Theta}] &= \int_{z_i^{({\ell})}} f(z_i^{({\ell})} | \tilde{s}_i^{(:{\ell}:)}, y_i, \hat{\Theta}) \int_{z_i^{({\ell}+1)}} f(z_i^{({\ell}+1)} | z_i^{({\ell})}, \tilde{s}_i^{(:k_{\ell}:)}, \hat{\Theta})[(z_i^{({\ell})}- \hat{\eta}_{k_{{\ell}}}^{({\ell})})z_i^{({\ell}+1)T}] dz_i^{({\ell}+1)} dz_i^{({\ell})} \\
    &\approx \sum_{m_{{\ell}}=1}^{M^{({\ell})}} f(z_{i, m_{{\ell}}}^{({\ell})} | \tilde{s}_i^{(:k_{\ell}:)}, y_i, \hat{\Theta}) \sum_{m_{{\ell}+1}=1}^{M^{({\ell}+1)}} [(z_{i, m_{\ell}}^{({\ell})} - \hat{\eta}_{k_{l}}^{({\ell})})z_{i, m_{{\ell}+1}}^{({\ell}+1)T}],
\end{align*}
\end{small}
where $z_{i, m_{\ell}}^{({\ell})}$ has been drawn from $f(z_{i,m_{{\ell}}}^{({\ell})} | s, \Hat{\Theta})$ and $z_{i, m_{{\ell}+1}}^{({\ell}+1)}$ from $f(z_{i,m_{{\ell}+1}}^{({\ell}+1)} | z_{i,m_{\ell}}^{({\ell})}, s, \Hat{\Theta})$.\\
Finally, we write
\small{

\begin{align*}
    &\mathbb{E}_{z^{({\ell})}, z^{({\ell}+1)}, s | y,  \hat{\Theta}}\left[\frac{\partial{\log L(z^{({\ell})} | z^{({\ell}+1)}, s, \Theta)}}{\partial{\Psi_{k_{{\ell}}}^{({\ell})}}}\right]
    = 0 \\
    &\iff -\frac{1}{2}\Psi^{({\ell})-1}_{k_{{\ell}}}\sum_{i=1}^n  \mathbb{E}_{z_i^{({\ell})}, z_i^{({\ell}+1)}, s_{i} | , y_i,  \hat{\Theta}}\Big[I_{r_1} - \Big{(}z_i^{({\ell})} - (\eta_{k_{{\ell}}}^{({\ell})} +  \Lambda_{k_{{\ell}}}^{({\ell})}z_i^{({\ell}+1)})\Big{)}\Big{(}z_i^{({\ell})} - (\eta_{k_{{\ell}}}^{({\ell})} +  \Lambda_{k_{{\ell}}}^{({\ell})}z_i^{({\ell}+1)})\Big{)}^T\Psi^{({\ell})-1}_{k_{{\ell}}}\Big]= 0 \\
    &\iff \sum_{i=1}^n \mathbb{E}_{z_i^{({\ell})}, z_i^{({\ell}+1)}, s_i| y_i,  \hat{\Theta}}\Big[I_{r_1}
    - e^{({\ell})}e^{({\ell})T}\Psi^{({\ell})-1}_{k_{{\ell}}}\Big]= 0 \\
    &\iff \sum_{i=1}^n \sum_{\tilde{s}_i^{(:k_{\ell}:)}} f(s_i^{(:k_{\ell}:)} = \tilde{s}_i^{(:k_{\ell}:)}| y_i, \hat{\Theta}) I_{r_1} = \sum_{i=1}^n \sum_{\tilde{s}_i^{(:k_{\ell}:)}} f(s_i^{(:k_{\ell}:)} = \tilde{s}_i^{(:k_{\ell}:)}| y_i, \hat{\Theta}) \mathbb{E}_{z_i^{({\ell})}, z_i^{({\ell}+1)} | \tilde{s}_i^{(:k_{\ell}:)} , y_i,  \hat{\Theta}}\Big[e^{({\ell})}e^{({\ell})T}\Big]\Psi^{({\ell})-1}_{k_{{\ell}}} = 0\\
    & \iff \sum_{i=1}^n \sum_{\tilde{s}_i^{(:k_{\ell}:)}} f(s_i^{(:k_{\ell}:)} = \tilde{s}_i^{(:k_{\ell}:)}| y_i, \hat{\Theta})\Psi^{({\ell})}_{k_{{\ell}}} = \sum_{i=1}^n \sum_{\tilde{s}_i^{(:k_{\ell}:)}} f(s_i^{(:k_{\ell}:)} = \tilde{s}_i^{(:k_{\ell}:)}| y_i, \hat{\Theta})\mathbb{E}_{z_i^{({\ell})}, z_i^{({\ell}+1)} |\tilde{s}_i^{(:k_{\ell}:)}, y_i,  \hat{\Theta}}\Big[e^{({\ell})}e^{({\ell})T}\Big], 
\end{align*}
}
with $e^{({\ell})} = \Big{(}z_i^{({\ell})} - (\eta_{k_{{\ell}}}^{({\ell})} +  \Lambda_{k_{{\ell}}}^{({\ell})}z_i^{({\ell}+1)})\Big{)}.$ 
Hence the estimator of $\Psi^{({\ell})}_{k_{{\ell}}}$ has the form
\[
\hat{\Psi}^{({\ell})}_{k_{{\ell}}} =
\frac{\sum_{i=1}^n \sum_{\tilde{s}_i^{(:k_{\ell}:)}} f(s_i^{(:k_{\ell}:)} = \tilde{s}_i^{(:k_{\ell}:)}| y_i, \hat{\Theta})E\Big[\Big{(}z_i^{({\ell})} - (\eta_{k_{{\ell}}}^{({\ell})} +  \Lambda_{k_{{\ell}}}^{({\ell})}z_i^{({\ell}+1)})\Big)\Big{(}z_i^{({\ell})} - (\eta_{k_{{\ell}}}^{({\ell})} +  \Lambda_{k_{{\ell}}}^{({\ell})}z_i^{({\ell}+1)})\Big)^T | \tilde{s}_i^{(:k_{\ell}:)}, y_i,  \hat{\Theta} \Big]}{\sum_{i=1}^n \sum_{\tilde{s}_i^{(:k_{\ell}:)}} f(s_i^{(:k_{\ell}:)} = \tilde{s}_i^{(:k_{\ell}:)}| y_i, \hat{\Theta})}.
\]

\subsection{Common tail layers mathematical derivations (E step)}

The conditional expectation $f(z^{(\ell)}, z^{(\ell+1)}, s^C, s^D, s^{({L_0}+1:)} | y^C, y^D,  \hat{\Theta}_C, \hat{\Theta}_D, \hat{\Theta}_{{L_0}+1:})$
can be rewritten as:
\begin{align}
     f(z^{(\ell)}, z^{(\ell+1)}, s^C, s^D, & s^{({L_0}+1:)} | y^C, y^D,  \hat{\Theta}_C, \hat{\Theta}_D, \hat{\Theta}_{{L_0}+1:}) \nonumber  \\
    &= f(z^{(\ell)} | s^C, s^D, s^{({L_0}+1:)},  y^C, y^D,  \hat{\Theta}_C, \hat{\Theta}_D, \hat{\Theta}_{{L_0}+1:}) 
    f(z^{(\ell+1)} | z^{(\ell)}, s^{({L_0}+1:)},  \hat{\Theta}_{{L_0}+1:}) \nonumber \\
    & ~~~~\times f(s^C, s^D, s^{({L_0}+1:)} |  y^C, y^D,  \hat{\Theta}_C, \hat{\Theta}_D, \hat{\Theta}_{{L_0}+1:}).
    \label{junc_layer_distrib}
\end{align}
$\forall{\ell} \in [{L_0} + 1, L]$, the first term of (\ref{junc_layer_distrib}) can be proportionally expressed as:
\begin{eqnarray*}
    f(z^{(\ell)} | s^C, s^D, s^{({L_0}+1:)},  y^C, y^D,  \hat{\Theta}_C, \hat{\Theta}_D, \hat{\Theta}_{{L_0}+1:}) &\propto& f(y^C | z^{(\ell)}, s^C, s^{({L_0}+1:)},  \hat{\Theta}_C, \hat{\Theta}_{{L_0}+1:})
    \\
    & &
    \times f(z^{(\ell)} | s^D, s^{({L_0}+1:)}, y^D, \hat{\Theta}_D, \hat{\Theta}_{{L_0}+1:}).
\end{eqnarray*}
One can compute $f(y^C | z^{(\ell)}, s^C, s^{({L_0}+1:)},  \hat{\Theta}_C, \hat{\Theta}_{{L_0}+1:})$ using Bayes rule and $f(z^{(\ell)} | s^D, s^{({L_0}+1:)}, y^D, \hat{\Theta}_D, \hat{\Theta}_{{L_0}+1:})$ is known from (\ref{rec_comput}).  
Finally the second term of (\ref{junc_layer_distrib}) can be computed as in (\ref{pz2_z1s}). By mutual independence of  $s^C, s^D$, and $ s^{({L_0}+1:)}$, the third term reduces to the product  of three densities which are given in  Section \ref{E-STEP-PATH}.

\subsection{Path probabilities mathematical derivations}
\subsubsection{E step for determining the path probabilities}
\label{E-STEP-PATH}
We consider the  three following densities:  $f(s^{({\ell} )D} = k_{\ell}  | y^D, \hat{\Theta}_D, \hat{\Theta}_{{L_0}+1:})$, $f(s^{({\ell} )C} = k_{\ell}  | y^C, \hat{\Theta}_C, \hat{\Theta}_{{L_0}+1:})$, and $f(s^{({\ell} )} = k_{\ell}  | y^C, y^D, \hat{\Theta}_C, \hat{\Theta}_D, \hat{\Theta}_{{L_0}+1:})$.
The first density can be computed from (\ref{ps_y}) as $$f(s^{({\ell} )D} = k_{\ell}  | y^D, \hat{\Theta}_D, \hat{\Theta}_{{L_0}+1:}) = \sum_{\tilde{s} \in \Omega^{(:k_{\ell} :)D}}f(s^{(1D:L)} = \tilde{s} | y^D, \hat{\Theta}_D, \hat{\Theta}_{{L_0} + 1:}),$$
where $\Omega^{(:k_{\ell} :)D}$ is the set of the full paths going through the component $k_{\ell} $ of layer ${\ell} $ of head $D$. 
The second density can be computed similarly using the fact that $f(y^C | s^C, s^{({L_0} + 1:)}, \hat{\Theta}_C, \hat{\Theta}_{{L_0} + 1:})$ is  Gaussian with  parameters $(\mu^{(1C:L)}, \Sigma^{(1C:L)})$. 
Concerning the last density,
we have to compute $p(s^{(L_0 + 1:)} | y^C, y^D, \hat{\Theta}_{D}, \hat{\Theta}_{C}, \hat{\Theta}_{(L_0+ 1:)})$.  
We are still making the two following conditional assumptions:
\begin{align*}
    (y^C \indep y^D) | z^{(L_0 + 1)}&{\rm \ \ and \ \ } (z^D \indep z^C) | z^{(L_0 + 1)},
    \end{align*}
     with $z^{(L_0 + 1)}$ the first common tail layer. We then have:

\begin{align*}
    &p(s^{(L_0 + 1:)} | y^C, y^D, \hat{\Theta}_{D},  \hat{\Theta}_{C}, \hat{\Theta}_{(L_0+ 1:)})\\
    &= \frac{p(s^{(L_0 + 1:)}, y^C, y^D | \hat{\Theta}_{D}, \hat{\Theta}_{C}, \hat{\Theta}_{(L_0+ 1:)})}{p(y^C, y^D)}\\
    &\propto p(s^{(L_0 + 1:)}, y^C, y^D | \hat{\Theta}_{D}, \hat{\Theta}_{C}, \hat{\Theta}_{(L_0+ 1:)})\\
    &= \sum_{s^C}\sum_{s^D}  \int_{z^{(L_0 + 1})} p(s^{(L_0 + 1:)}, y^C, y^D, z^{(L_0 + 1)}, s^C, s^D | \hat{\Theta}_{D}, \hat{\Theta}_{C}, \hat{\Theta}_{(L_0+ 1:)}) dz^{(L_0 + 1)}\\
    &= \sum_{s^C}\sum_{s^D} \int_{z^{(L_0 + 1})} p(y^C | s^C, s^{(L_0 + 1:)}, z^{(L_0 + 1)}, \hat{\Theta}_{C}, \hat{\Theta}_{(L_0+ 1:)}) p(y^D | s^D, s^{(L_0 + 1:)}, z^{(L_0 + 1)}, \hat{\Theta}_{D}, \hat{\Theta}_{(L_0+ 1:)})\\
    &\times p(z^{(L_0 + 1)} |  s^{(L_0 + 1:)},  \hat{\Theta}_{(L_0+ 1:)}) \prod_{{\ell} =1}^{L_0}p(s^{({\ell})C} | \hat{\Theta}_{C}, \hat{\Theta}_{(L_0+ 1:)})p(s^{({\ell})D} | \hat{\Theta}_{D}, \hat{\Theta}_{(L_0+ 1:)})\prod_{{\ell} = L_0 + 1}^{L}  p(s^{({\ell})} | \hat{\Theta}_{(L_0+ 1:)})dz^{(L_0 + 1)},
\end{align*}%
by independence of the $(s^{(\ell)h})_{\ell,h}$. The first two terms are computed as for (\ref{junc_layer_distrib}), the third term is a Gaussian given in (\ref{gauss_distrib}) and each density of the products are multinomial densities whom coefficients have already been estimated.

%

\subsubsection{M step for determining the path probabilities}
%
Using again the conditional  independence, we maximise
\begin{align*}
    E&_{s^C, s^D, s^{({L_0}+1)}|y^C, y^D, \hat{\Theta}_C, \hat{\Theta}_D, \hat{\Theta}_{{L_0}+1:}}[\log L(s^C, s^D, s^{({L_0} + 1:)} | \Theta_C, \Theta_D, \Theta_{{L_0} + 1:})] \\
    &= \sum_{{\ell} =1}^{L_0} \mathbb{E}_{s^{({\ell} )C}  |y^C, \hat{\Theta}_C, \hat{\Theta}_{{L_0}+1:}}[\log L(s^{(l)C}  | \Theta_C, \Theta_{{L_0} + 1:})]\\
    &+ \sum_{{\ell} =1}^{L_0} \mathbb{E}_{s^{({\ell} )D} | y^D, \hat{\Theta}_D, \hat{\Theta}_{{L_0}+1:}}[\log L(s^{(l)D} | \Theta_D, \Theta_{{L_0} + 1:})]\\
    &+ \sum_{{\ell} ={L_0} + 1}^{L} \mathbb{E}_{s^{({\ell} )}|y^C, y^D, \hat{\Theta}_C, \hat{\Theta}_D, \hat{\Theta}_{{L_0}+1:}}[\log L(s^{({\ell} )} | \Theta_C, \Theta_D, \Theta_{{L_0} + 1:})],
\end{align*}
with respect to $\pi_{k_{\ell} }^{({\ell} )h}$, $\forall{h \in \{C,D\}}, \ell \in [1,{L_0}], k_{\ell} \in [1,K_{\ell} ]$ and  with respect to $\pi_{k_{\ell} }^{({\ell} )}$, $\forall{{\ell} } \in [{L_0}, L], k_{\ell} \in [1,K_{\ell} ]$.

Each of heads and tail estimators can be computed in the same way. Let $k_{\ell}$ be the index of a component of layer ${\ell}$ for which we want to derive an estimator and $\Tilde{k}_{\ell}$ another component index. The associated probabilities are respectively $\pi_{k_{\ell}}$ and  $\pi_{\Tilde{k}_{\ell}}$. We omit the head subscript $h$ for better readability.\\
We have
\begin{align*}
    E&_{s^{({\ell})}|y, \hat{\Theta}}[\log L(s^{({\ell})} | \Theta)] \\
    &= \sum_{i = 1}^n\sum_{k_{\ell}' = 1}^{K_{\ell}} f(s^{({\ell})} = k_{\ell}' | y, \hat{\Theta}) \log L(s^{({\ell})} = k_{\ell}' | \Theta) \\
    &= \sum_{i = 1}^n \sum_{\substack{k_{\ell}' = 1 \\ k_{\ell}'\neq \Tilde{k}_{\ell}}}^{K_{\ell}} f(s^{({\ell})} = k_{\ell}' | y, \hat{\Theta}) \log L(s^{({\ell})} = k_{\ell}' | \Theta) + \sum_{i = 1}^n f(s^{({\ell})} = \Tilde{k}_{\ell} | y, \hat{\Theta}) \log L(s^{({\ell})} = \Tilde{k}_{\ell} | \Theta) \\
    &= \sum_{i = 1}^n \sum_{\substack{k_{\ell}' = 1 \\ k_{\ell}'\neq \Tilde{k}_{\ell}}}^{K_{\ell}} f(s^{({\ell})} = k_{\ell}' | y, \hat{\Theta}) \log \pi_{k_{\ell}'}^{({\ell})} + \sum_{i = 1}^n f(s^{({\ell})} = \Tilde{k}_{\ell} | y, \hat{\Theta}) \log(1 - \sum_{\substack{k_{\ell}' = 1 \\ k_{\ell}'\neq \Tilde{k}_{\ell}}} \pi_{k_{\ell}'}^{({\ell})}).
\end{align*}
Taking the derivative with respect to $\pi_{k_{\ell}}^{({\ell})}$ and equalizing to zero yields

\begin{align*}
    \frac{\partial \mathbb{E}_{s^{({\ell})}|y, \hat{\Theta}}[\log L(s^{({\ell})} | \Theta)]}{{\partial \pi_{k_{\ell}}^{({\ell})}}} = 0
    & \Leftrightarrow \frac{\sum_{i = 1}^n  f(s^{({\ell})} = k_{\ell} | y, \hat{\Theta})}{\pi_{k_{\ell}}^{({\ell})}} = \frac{\sum_{i = 1}^n f(s^{({\ell})} = \Tilde{k}_{\ell} | y, \hat{\Theta})}{\pi_{\Tilde{k}_\ell}^{({\ell})}}\\
    & \Leftrightarrow \pi_{\Tilde{k}_{\ell}}^{({\ell})} = \frac{\sum_{i = 1}^n f(s^{({\ell})} = \Tilde{k}_{\ell} | y, \hat{\Theta})}{\sum_{i = 1}^n  f(s^{({\ell})} = k_{\ell} | y, \hat{\Theta})}\pi_{k_{\ell}}^{({\ell})}. 
\end{align*}
Finally, summing over $\Tilde{k}_{\ell}$ we get
\begin{align*}
    \pi_{\Tilde{k}_{\ell}}^{({\ell})} = \frac{\sum_{i = 1}^n f(s^{({\ell})} = \Tilde{k}_{\ell} | y, \hat{\Theta})}{\sum_{i = 1}^n  f(s^{({\ell})} = k_{\ell} | y, \hat{\Theta})}\pi_{k_{\ell}}^{({\ell})}
     \Leftrightarrow 1 = \frac{n}{\sum_{i = 1}^n  f(s^{({\ell})} = k_{\ell} | y, \hat{\Theta})}\pi_{k_{\ell}}^{({\ell})}
     \Leftrightarrow \hat{\pi}_{k_{\ell}}^{({\ell})} = \frac{\sum_{i = 1}^n  f(s^{({\ell})} = k_{\ell} | y, \hat{\Theta})}{n}. 
\end{align*}\\
As a result, the probability estimator for each head $h$ is:
$$ \hat{\pi}_{k_{\ell} }^{({\ell} )h} = \frac{\sum_{i = 1}^n  f(s^{({\ell} )h} = k_{\ell}  | y^h, \hat{\Theta}_h, \hat{\Theta}_{{L_0}+1:})}{n}.$$
For the common tail the estimator is of the form, $\forall {{\ell} } \in [{L_0} + 1,L]$:
$$ \hat{\pi}_{k_{\ell} }^{({\ell} )} = \frac{\sum_{i = 1}^n  f(s^{({\ell} )} = k_{\ell}  | y^C, y^D, \hat{\Theta}_C, \hat{\Theta}_D, \hat{\Theta}_{{L_0}+1:})}{n}.$$

\subsection{Latent variables identifiabiliy rescaling}\label{gllvm_rescaling}

The GLLVM and Factor Analysis models assume that the latent variable is centered and of unit variance, \textit{i.e.} that $z^{(1)C}$ and $z^{(1)D}$ are centered-reduced in our setup.\\
We iteratively center and reduce each layer latent variable $z^{(\ell)}$ starting from the last layer of the common tail to the first layers of each head $h$ in order for all $(z^{(\ell)h})_{h,\ell}$ to be centered-reduced. As the latent variable of the last layer of DGMM family models is a centered-reduced Gaussian, by induction, $z^{(1)C}$ and $z^{(1)D}$ are centered and reduced.\\

Assuming that the latent variable of the next layer is centered-reduced, the mean and variance of the latent variable of the current layer $l \in [1,L]$ of head or tail $h \in \{C, D, L_0 + 1:\}$ is:

$$\begin{cases}
    E(z^{(\ell)h}) = \sum_{k'_{\ell}} \pi_{k'_{\ell}}^{(\ell)h} \eta_{k'_{\ell}}^{(\ell)h} \\
    Var(z^{(\ell)h}) = \sum_{k'_{\ell}} \pi_{k'_{\ell}}^{(\ell)h}(\Lambda_{k'_{\ell}}^{(\ell)h}  \Lambda_{k'_{\ell}}^{(\ell)hT} + \Psi_{k'_{\ell}}^{(\ell)h} +  \eta_{k'_{\ell}}^{(\ell)h} \eta_{k'_{\ell}}^{(\ell)hT}) - (\sum_{k'_{\ell}} \pi_{k'_{\ell}}^{(\ell)h} \eta_{k'_{\ell}}^{(\ell)h})(\sum_{k'_{\ell}} \pi_{k'_{\ell}}^{(\ell)h} \eta_{k'_{\ell}}^{(\ell)h})^T.
\end{cases}$$
Let $A^{(\ell)h}$ be the Cholesky decomposition of $Var(z^{(\ell)h})$ $\forall{k_{\ell}} \in [1, K_{\ell}]$, then we rescale the layer parameters in the following way:

$$\begin{cases}
    \eta_{k_\ell}^{(\ell)h new} = A^{(l)h-1T}\left[\eta_{k_\ell}^{(\ell)h} - \sum_{k'_{\ell}} \pi_{k'_{\ell}}^{(\ell)h} \eta_{k'_{\ell}}^{(\ell)h} \right] \\
    \Lambda_{k_\ell}^{(\ell)h new} = A^{(l)h-1T}\Lambda_{k_\ell}^{(\ell)h} \\
    \Psi_{k_\ell}^{(\ell)new} = A^{(l)h-1T}\Psi_{k_\ell}^{(\ell)h}A^{(l)h-1},
\end{cases}$$
with the subscript ``new" denoting the rescaled version of the parameters.

\subsection{Monte Carlo scheme}
\label{MCS}
The number of Monte Carlo copies $M^{(\ell)h}$ to draw at each layer has to be chosen before running the MCEM.
\cite{wei1990monte} advise to let $M$ grow through the iterations starting with a very low $M$. Doing so, one does not get stuck into very local optima at the beginning of the algorithm and ends up in a precisely estimated expectation state. 
The growth scheme of $M_t^{\ell}$ through the iterations $t$ implemented here is:
$$M_t^{\ell} = \left\lfloor\frac{40}{\log(n)} \times t \times \sqrt{r_{\ell}}\right\rfloor.$$
$M_t^{\ell}$ grows linearly with the number of iterations $t$ to follow \cite{wei1990monte} advice. In order to explore the latent space at each layer, $M^{\ell}$ also grows with the dimension of the layer. The square root rate just ensures that the running time remains affordable, whereas if there were no additional computational costs we would certainly have let $M^{\ell}$ grow much more with $r_{\ell}$. Finally, we make the hypothesis that the more observations in the dataset the stronger the signal is and hence the fewer draws of latent variables are needed to train the model.

\begin{rmk}
In this Monte Carlo version contrary to the regular EM algorithm, the likelihood does not increase necessary  through the iterations. In classical EM-based models, the training is often stopped once the likelihood increases by less than a given threshold between two iterations. The stopping process had then to be adapted to account for temporary losses in likelihood. Hence, we have defined a patience parameter which is the number of iterations without log-likelihood increases to wait before stopping the algorithm. Typically, we set this parameter to 1 iteration in the simulations.
\end{rmk}%

\subsection{Model selection details}\label{model_selection}
This section gives additional details about the way model selection is performed on the fly. 

A component of the $\ell$th layer is considered useless if its probability is inferior to $\frac{1}{4 k_{\ell}}$, where $k_{\ell}$ denotes the number of components of the layer. For instance, if a layer is formerly composed of four components, the components associated with a probability inferior to 0.0625 are removed from the architecture. 

For the GLLVM layer, logistic and linear regressions were fitted to determine which of the dimensions had a significant effect over each $y_j^D$ for each path $\tilde{s}$. We have fitted a logistic LASSO for each binary, count and categorical variable and an ordinal logistic regression for each ordinal variable. In the M1DGMM case, we have fitted a linear LASSO for each continuous variable. 
The variables associated with coefficients identified as being zero (or not significant at a 10\% level) for at least 25\% of the paths were removed. 

The same voting idea was used for the regular DGMM layers to determine the useless dimensions. As our algorithm generate draws of $(z^{({\ell}+1)} | z^{({\ell})}, s)$ of dimension $r_{{\ell} + 1}$, it is possible to perform a PCA on this variable for each path and each of the $M^{({\ell})}$ points simulated for $z^{({\ell})}$.
Doing so, one can compute the average contribution of each dimension of $r_{l+1}$ to the first principal component and set a threshold under which a dimension is deleted. We have set this threshold to $0.2$ for our simulations. The intuition behind this is that the first component of the PCA conveys the majority of the pieces of information that $z^{({\ell}+1)}$ has on $z^{({\ell})}$. If a dimension shares no common information with this first component, hence it is not useful to keep it.

The dimension of the junction layer (the first DGMM layer on the common tail) is chosen according to this procedure too. The two heads decide which dimensions of the junction layer is important and each dimension important for at least one head is kept. This is rather conservative but avoids that contradictory information coming from the two heads disrupt the global architecture. 

The number of layers on the heads and tails is fully determined by the selection of the layers dimensions in order to keep the model identifiable. If the dimension of an intermediate tail layer ${\ell}$ is selected to be one then $r_{\ell} > r_{{\ell} + 1} > ... > r_L$ does not hold anymore. Thus, the following tail layers are deleted. \\
Similarly, if an head layer has a selected dimension of two, then the following head layers are deleted. Indeed, the tail has to have minimal dimensions of two and one on its last layers. This is not compatible with previous head layers of dimension inferior or equal to two. 

In the case of head layers deletion, we restart the algorithm (initialisation and proper model run) with the new architecture. Otherwise it would be necessary to re-determine all the path and DGMM coefficients values to bridge the gap between the previous head layer and the junction layer. There were no easy way to do such thing and restarting the algorithm seemed the best to do. Note that in our simulations defining several heads layers did not give good results. Intuitively, it could too much dilute information before passing it to the common tail, resulting in poor performance. We advise to keep only one or two head layers before running the MDGMM. Doing so, this restarting procedure would not be often performed in practice.

\subsection{Metrics}\label{metrics}
A true positive (TP) prediction of the model is an observation that has been assigned to the same class as the ``ground truth" label. On the contrary, a False Positive (FP) means that the class predicted by the model and the label do not match. $k$ denotes the class index and $K$ the cardinal of the set of all possible classes. $n_k$ is the number of points in the class $k$ and $y_{i,k}$ an observation of class $k$.\\

The formulas of the two precision metrics are :
\begin{align*}
    \text{Micro precision} &= \frac{\sum_{k = 1}^K\sum_{i = 1}^n TP_{i,k}}{\sum_{k = 1}^K\sum_{i = 1}^n TP_{i,k} + FP_{i,k}},\\
    \text{Macro precision} &= \frac{1}{K}\sum_{k = 1}^K\frac{\sum_{i = 1}^n TP_{i,k}}{\sum_{i = 1}^n TP_{i,k} + FP_{i,k}}.\\
\end{align*}

The formula of the silhouette coefficient is:
\[
    \text{Silhouette coefficient} = \frac{1}{K}\sum_{k = 1}^K\frac{1}{n_k}\sum_{i =1}^n\frac{d\_inter(i, k) - d\_intra(i, k)}{\max{(d\_intra(i, k), d\_inter(i, k)})},
\]

with $d\_intra(i, k) = \frac{1}{n_k - 1}\sum_{i' \neq i} d(y_{i,k}, y_{i',k})$ and $d\_inter(i, k) = \min_{k'\neq k}\frac{1}{n_k}\sum_{i' \neq i}\sum_{k'=1}^K d(y_{i,k}, y_{i',k'})$

With $d$ a distance, the Gower distance \citep{gower1971general} in our case.

\subsection{Benchmark models specifications}\label{model_spe}
A standard Grid Search has been performed to find the best specification of the hyperparameters of the benchmark models. The best value for each metric is reported independently from the other metrics. As such for a given model, the best silhouette score, micro and macro precisions can actually be achieved by three different specifications. The silhouette metric seemed to us the most appropriate since it is unsupervised, but we did not want to favor any
metric against the others. Besides, all of the benchmark models are not built upon a likelihood principle which prevents from performing model selection using a common criterion such as the Bayesian Information Criterion (BIC) or the Aikake Information Criterion (AIC). Therefore, this performance report aims at illustrating the clustering power of different algorithms compared to the ones introduced in this work rather than presenting the metrics associated with the best selected specification of each benchmark model.\\

The following hyperparameters search spaces were used : \\

\textbf{K-modes} (from the kmodes package)
\begin{itemize}
    \item Initialisation $ \in$  \{'Huang', 'Cao', 'random'\}.
\end{itemize}

\vspace{2mm}

\textbf{K-prototypes} (from the kmodes package)
\begin{itemize}
    \item Initialisation $ \in$  \{'Huang', 'Cao', 'random'\}.
\end{itemize}

\vspace{2mm}

\textbf{Agglomerative clustering} (from the scikit-learn package)
\begin{itemize}
    \item linkages $ \in$ \{'complete', 'average', 'single'\}.
\end{itemize}
This model was trained using the Gower Distance Matrix computed on the data.
\vspace{2mm}

\textbf{Self-Organizing Map} (from the SOMPY package)
\begin{itemize}
    \item sigma $ \in$ [0.001, 0.751, 1.501, 2.250, 3.000]
    \item lr $ \in$ [0.001, 0.056, 0.111, 0.167, 0.223, 0.278, 0.333, 0.389, 0.444, 0.500].
\end{itemize}

\vspace{2mm}

\textbf{DBSCAN} (from the scikit-learn package)
\begin{itemize}
    \item leaf\_size  $ \in$ \{10, 20, 30, 40, 50\}
    \item eps $ \in$ \{0.01, 1.258, 2.505, 3.753, 5.000\}
    \item min\_samples $\in$ \{1, 2, 3, 4\}
    \item Data used: {'scaled data', 'Gower Distance'}.
\end{itemize}

DBSCAN was trained on two versions of the dataset: on the data themselves and using the Gower Distance Matrix computed on the data. Each time the best performing specification was taken. \\

\vspace{2mm}

\textbf{GLMLVM}
\begin{itemize}
    \item $r \in [1, 5]$
    \item $k = 2$.
\end{itemize}

\vspace{2mm}

\textbf{NESP DDGMM (MCA + GMM + FA)}
\begin{itemize}
    \item $r \in [1, 13]$
    \item $k = 2$.
\end{itemize}

\vspace{2mm}

\textbf{DDGMM}\\
The starting architecture over which automatic architecture selection was performed was:
\begin{itemize}
    \item $r = \{5, 4, 3 \}$
    \item $k = \{4, 2\}$
    \item Number of maximum iterations $= 30$.
\end{itemize}

\vspace{2mm}

\textbf{NESP M2DGMM (MCA + GMM + FA + PLS)}\\
The architectures considered had at most 2 layers on each head and 3 layers on the tail.
\begin{itemize}
    \item $r$: All the minimal identifiable architectures.
    \item $k$: Random draws for each $k_{\ell}$ $\in$ \{2, 3, 4\}
    \item Number of maximum iterations $= 30$.

\end{itemize}

\vspace{2mm}

\textbf{M1DGMM}\\
The starting architecture over which automatic architecture selection was performed was:
\begin{itemize}
    \item $r = \{5, 4, 3\}$
    \item $k = \{4, 2\}$
    \item Number of maximum iterations $= 30$.
\end{itemize}

\vspace{2mm}
\textbf{M2DGMM}\\
The starting architecture over which automatic architecture selection was performed was:
\begin{itemize}
    \item $r_c = \{p_c\}, r_d = \{5\}, r_t = \{4, 3\}$
    \item $k_c = \{1\}, k_d = \{3\}, k_{L_0 + 1:} = \{2, 1\}$
    \item Number of maximum iterations $= 30$.
\end{itemize}

$k_c = \{1\}$ and $r_c = \{p_c\}$ are imposed by construction as the first layer of the continuous head are the data themselves.\\

\end{document}